\documentclass{article} 
\usepackage[preprint]{colm2026_conference}

\usepackage{microtype}
\usepackage{hyperref}
\usepackage{url}
\usepackage{booktabs}
\usepackage{graphicx}
\usepackage{verbatim}
\usepackage{algorithm}
\usepackage{algorithmic}
\usepackage{amsmath}
\usepackage{amsthm}
\usepackage{nicefrac}       
\usepackage{amsfonts}       
\usepackage{tikz}
\usetikzlibrary{positioning, calc, shapes.geometric, decorations.pathreplacing, arrows.meta, fit}
\usepackage{enumitem}
\usepackage{subcaption}
\usepackage{siunitx}
\usepackage[T1]{fontenc}

\usepackage[final, commandnameprefix=ifneeded]{changes}

\usepackage{lineno}

\definecolor{darkblue}{rgb}{0, 0, 0.5}
\hypersetup{colorlinks=true, citecolor=darkblue, linkcolor=darkblue, urlcolor=darkblue}

\title{KDFP: A first-principles approach to knowledge distillation in large language models}

\author{Ryan Swift \& Konstantinos Psounis 
\\
Thomas Lord Department of Computer Science\\
University of Southern California\\
Los Angeles, CA 90007, USA \\
\texttt{\{ryanswif,kpsounis\}@usc.edu}
}

\begin{document}

\ifcolmsubmission
\linenumbers
\fi

\maketitle

\begin{abstract}
Knowledge distillation is an established technique for improving the capabilities of small, efficient student models by training them with the representations of larger, more capable teacher models. Much of the recent work in the distillation of large language models (LLMs) has focused on distilling abilities learned during post-training, such as instruction following, chain-of-thought reasoning, and tool usage. This has left a large research gap in general knowledge distillation for LLMs, which is essential for developing efficient and private systems suitable for deployment on edge devices. We take a first-principles approach, evaluating previous lessons from prior works and conducting new explorations to develop a distillation methodology suitable for modern LLMs. We present KDFP, a novel methodology for white-box general knowledge distillation in LLMs. We demonstrate that KDFP outperforms existing methods by 1.6\% -- 4.9\% across 9 benchmarks while increasing training efficiency by up to 99.1\% through ephemeral parameter reduction.
\end{abstract}

\section{Introduction}
\label{intro}
The machine learning landscape has undergone radical changes since knowledge distillation (KD) was introduced by \citet{bucilua_model_2006} and formulated for neural networks by \citet{hinton_distilling_2015}. The current era is dominated by the large language models (LLMs) and the transformer architecture \citep{vaswani_attention_2017}, with parameter counts in the hundreds of billions \citep{meta_ai_llama_2025, team_kimi_2025, grattafiori_llama_2024}. 

Skills such as instruction-following, reasoning, and tool usage have been critical to improving the practical utility of LLMs in applications such as agentic systems. As such, much of the recent work in KD has been focused on distilling these particular abilities \citep{yang_survey_2024, zhu_survey_2024, xu_survey_2024}. Comparatively less work has been done to advance general knowledge distillation. 

Despite the gap in research activity, it has never been clearer just how important general KD is to the advancement of LLMs. LLMs can be compressed with KD to create highly capable and privacy-preserving on-device systems for resource-constrained platforms such as smartphones and single-board computers. Distilled LLMs small enough to run on a workstation have surpassed other models with 15x as many parameters \cite{team_gemma_2025}.

We draw upon the lessons of previous works while also exploring novel formulations to determine how general KD can be improved. We work to unite the findings of previous works such as TED \citep{liang_less_2023} and TinyBERT \citep{jiao_tinybert_2020} with what we now know about the science of LLMs from the findings of mechanistic interpretability \citep{elhage_mathematical_2021, elhage_toy_2022, lindsey_sparse_2024}.

The core contribution of this paper is a novel methodology for general knowledge distillation in transformers, which we refer to as Knowledge Distillation from First Principles (KDFP). We compare KDFP against TED \citep{liang_less_2023}-- the most recent work in white-box general KD \citep{xu_survey_2024, zhu_survey_2024}-- as well as earlier methods previously surpassed by TED. In Section \ref{ss:main_results} we demonstrate that KDFP outperforms existing methods by 1.6\% -- 4.9\% across 9 benchmarks, and Section \ref{ss:efficiency} details an efficiency increase of up to 99.1\%. In Section \ref{ss:explaining_ted} we leverage an understanding of the internal mechanisms of LLMs to provide a possible explanation for why TED fails to generalize to modern models and evaluations.

Beyond the KDFP itself, we present 4 additional contributions:
\begin{enumerate}
    \item Section \ref{ss:main_results} demonstrates that previously ignored activations are essential for peak student performance
    \item Sections \ref{ss:ft_teacher} and \ref{ss:big_teacher} illustrate how students may benefit from teachers whose representations are more closely aligned with their own
    \item Section \ref{ss:5x_data} shows that increasing training data may decrease student performance
    \item Section \ref{s:results} indicates that a target distribution mix of 75\% teacher and 25\% student outperforms existing divergence formulations
\end{enumerate}

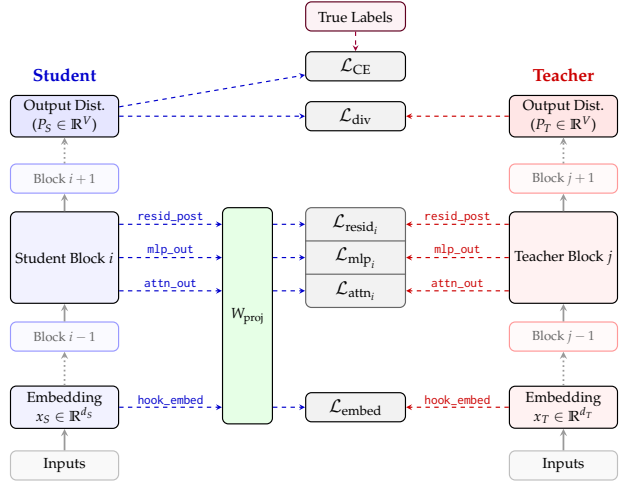
\begin{figure}[tbp]
\centering
\scalebox{0.55}{
\begin{tikzpicture}[
    >=stealth,
    thick,
    box/.style={draw, rounded corners, align=center, minimum height=0.7cm},
    student/.style={box, fill=blue!5, minimum width=2.6cm},
    teacher/.style={box, fill=red!5, minimum width=2.6cm},
    s_generic/.style={box, fill=blue!2, draw=blue!40, minimum width=2.6cm, font=\small, text=gray!80!black},
    t_generic/.style={box, fill=red!2, draw=red!40, minimum width=2.6cm, font=\small, text=gray!80!black},
    proj/.style={box, fill=green!10, minimum width=1.0cm},
    loss/.style={box, fill=gray!10, inner sep=4pt, font=\large, minimum width=2.4cm},
    forward/.style={->, very thick, color=gray!80}, 
    distill/.style={->, dashed}
]

\def\sx{0}      
\def\px{4.4}    
\def\lx{7}   
\def\tx{12.0}   

\def\yinputs{-1.4}
\def\yemb{0.0}
\def\yprev{1.7}
\def\yblock{3.6} 
\def\ynext{5.5}
\def\ydist{7.0}  
\def\yce{8.2}    
\def\ytrue{9.4}  

\node[font=\large\bfseries, text=blue!80!black] at (\sx, \ydist + 1.0) {Student};
\node[font=\large\bfseries, text=red!80!black] at (\tx, \ydist + 1.0) {Teacher};

\node[box, fill=gray!5, draw=gray!50, minimum width=2.6cm] (s_in) at (\sx, \yinputs) {Inputs};
\node[box, fill=gray!5, draw=gray!50, minimum width=2.6cm] (t_in) at (\tx, \yinputs) {Inputs};

\node[student]   (s_emb)    at (\sx, \yemb)   {Embedding \\ $x_S \in \mathbb{R}^{d_S}$};
\node[s_generic] (s_prev)   at (\sx, \yprev)  {Block $i-1$};
\node[student, minimum height=2.2cm] (s_block) at (\sx, \yblock) {Student Block $i$};
\node[s_generic] (s_next)   at (\sx, \ynext)  {Block $i+1$};
\node[student, fill=blue!10] (s_dist) at (\sx, \ydist) {Output Dist. \\ ($P_S \in \mathbb{R}^{V}$)};

\draw[forward] (s_in) -- (s_emb);
\draw[forward, dotted] (s_emb) -- (s_prev);
\draw[forward] (s_prev) -- (s_block);
\draw[forward] (s_block) -- (s_next);
\draw[forward, dotted] (s_next) -- (s_dist);

\node[teacher]   (t_emb)    at (\tx, \yemb)   {Embedding \\ $x_T \in \mathbb{R}^{d_T}$};
\node[t_generic] (t_prev)   at (\tx, \yprev)  {Block $j-1$};
\node[teacher, minimum height=2.2cm] (t_block) at (\tx, \yblock) {Teacher Block $j$};
\node[t_generic] (t_next)   at (\tx, \ynext)  {Block $j+1$};
\node[teacher, fill=red!10] (t_dist) at (\tx, \ydist) {Output Dist. \\ ($P_T \in \mathbb{R}^{V}$)};

\draw[forward] (t_in) -- (t_emb);
\draw[forward, dotted] (t_emb) -- (t_prev);
\draw[forward] (t_prev) -- (t_block);
\draw[forward] (t_block) -- (t_next);
\draw[forward, dotted] (t_next) -- (t_dist);


\node[proj, minimum height=5.2cm, minimum width=1.2cm] (p_shared) at (\px, 2.2) {$W_{\text{proj}}$};

\node[loss] (l_emb) at (\lx, \yemb) {$\mathcal{L}_{\text{embed}}$};
\draw[distill, blue!80!black] (s_emb) -- (p_shared.west |- s_emb) node[midway, above=-2pt, font=\small] {\texttt{hook\_embed}};
\draw[distill, blue!80!black] (p_shared.east |- l_emb) -- (l_emb);
\draw[distill, red!80!black]  (t_emb) -- (l_emb) node[midway, above=-2pt, font=\small] {\texttt{hook\_embed}};

\node[box, fill=gray!10, draw=gray!80!black, minimum height=2.4cm, minimum width=2.4cm, inner sep=0pt] (l_block) at (\lx, \yblock) {};
\draw[gray!80!black] ([yshift=0.4cm]l_block.west) -- ([yshift=0.4cm]l_block.east);
\draw[gray!80!black] ([yshift=-0.4cm]l_block.west) -- ([yshift=-0.4cm]l_block.east);
\node[font=\large] at (\lx, \yblock + 0.8) {$\mathcal{L}_{\text{resid}_i}$};
\node[font=\large] at (\lx, \yblock) {$\mathcal{L}_{\text{mlp}_i}$};
\node[font=\large] at (\lx, \yblock - 0.8) {$\mathcal{L}_{\text{attn}_i}$};

\coordinate (s_resid) at ([yshift=0.8cm]s_block.east);
\coordinate (s_mlp)   at (s_block.east);
\coordinate (s_attn)  at ([yshift=-0.8cm]s_block.east);

\coordinate (l_resid) at ([yshift=0.8cm]l_block.west);
\coordinate (l_mlp)   at (l_block.west);
\coordinate (l_attn)  at ([yshift=-0.8cm]l_block.west);

\draw[distill, blue!80!black] (s_resid) -- (p_shared.west |- s_resid) node[midway, above=-2pt, font=\small] {\texttt{resid\_post}};
\draw[distill, blue!80!black] (s_mlp) -- (p_shared.west |- s_mlp) node[midway, above=-2pt, font=\small] {\texttt{mlp\_out}};
\draw[distill, blue!80!black] (s_attn) -- (p_shared.west |- s_attn) node[midway, above=-2pt, font=\small] {\texttt{attn\_out}};

\draw[distill, blue!80!black] (p_shared.east |- l_resid) -- (l_resid);
\draw[distill, blue!80!black] (p_shared.east |- l_mlp) -- (l_mlp);
\draw[distill, blue!80!black] (p_shared.east |- l_attn) -- (l_attn);

\draw[distill, red!80!black] ([yshift=0.8cm]t_block.west) -- ([yshift=0.8cm]l_block.east) node[midway, above=-2pt, font=\small] {\texttt{resid\_post}};
\draw[distill, red!80!black] (t_block.west) -- (l_block.east) node[midway, above=-2pt, font=\small] {\texttt{mlp\_out}};
\draw[distill, red!80!black] ([yshift=-0.8cm]t_block.west) -- ([yshift=-0.8cm]l_block.east) node[midway, above=-2pt, font=\small] {\texttt{attn\_out}};

\node[loss] (l_kd) at (\lx, \ydist) {$\mathcal{L}_{\text{div}}$};
\draw[distill, blue!80!black] (s_dist) -- (l_kd);
\draw[distill, red!80!black]  (t_dist) -- (l_kd);

\node[loss] (l_ce) at (\lx, \yce) {$\mathcal{L}_{\text{CE}}$};
\node[box, fill=purple!5, draw=purple!50!black, minimum width=2.4cm] (true_labels) at (\lx, \ytrue) {True Labels};

\draw[distill, blue!80!black] (s_dist) -- (l_ce);
\draw[distill, purple!80!black] (true_labels) -- (l_ce);

\end{tikzpicture}
}
\caption{Diagram of KDFP. The attention, MLP, and residual outputs are distilled in each block, as well as embeddings and model outputs. A single projection matrix is learned during training to project student activations into the teacher's embedding space.}
\label{fig:kdfp}
\end{figure}

\section{Preliminaries}
\label{prelim}

\subsection{Knowledge Distillation}
Knowledge distillation (KD) was introduced by \citet{bucilua_model_2006} as a technique for compressing the knowledge of a large, highly capable \textit{teacher} model into a smaller, more efficient \textit{student} model. The original formulation worked by generating synthetic data from teachers and using it to train student models for classification. 

Later work by \citet{hinton_distilling_2015} used a different approach to KD by formulating the problem as distribution matching, where students learn to approximate the output distribution of the teacher. Students were trained using forward Kullback-Leibler (KL) divergence on the output distributions produced by the teacher model.

\subsubsection{White-Box vs Black-Box}
\label{sss:whitevblack}
Black-box KD methods are defined by their sole reliance on inputs and outputs to train student models. This is the only option when the teacher model is proprietary \citep{openai_text_2025, anthropic_get_2025, google_gemini_2025}. Black-box methods such as synthetic data generation have proven to be very effective, producing instruction-tuned LLMs such as Vicuna \citep{zheng_judging_2023}, and Alpaca \citep{taori_alpaca_2023}.

In contrast, white-box KD methods make use of the information-rich internal activations. The traditional KD method proposed by \citet{hinton_distilling_2015} only distills output distributions, while recent methods for transformers also distill residual stream activations \citep{jiao_tinybert_2020, liang_less_2023}.

\subsection{Divergence Functions}
\label{div}
Divergence functions measure the \textit{statistical distance} between a true distribution $P$ and its approximation $Q$. The value of a divergence function is always nonnegative, with zero indicating a perfect match. Divergence functions are preferred over cross-entropy in distillation settings precisely for these properties.

\begin{figure}[tb]
    \centering
    \includegraphics[width=0.6\textwidth]{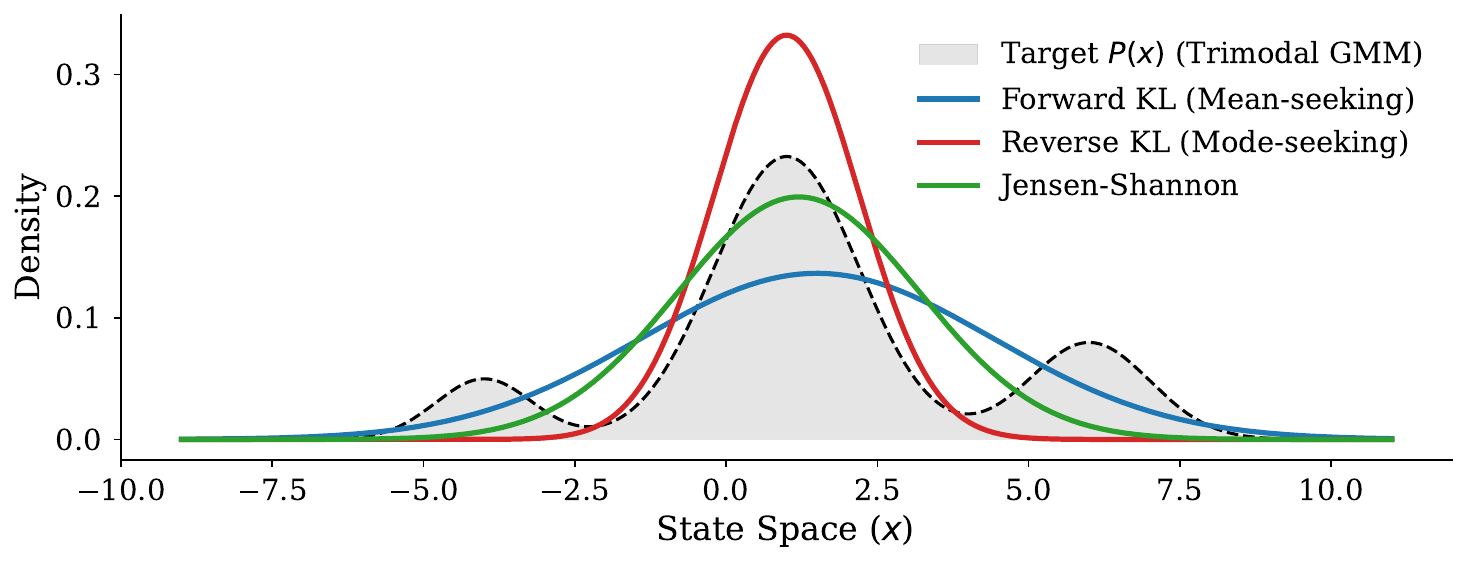}
    \caption{Comparison of JS divergence with Forward and Reverse KL divergence matching a Gaussian to a trimodal Gaussian mixture model (GMM).}
    \label{fig:jsd_kld_plot}
\end{figure}

\subsubsection{Kullback-Leibler Divergence}
\paragraph{Forward KL Divergence} Divergence was formulated for use in the distillation of neural networks by \citet{hinton_distilling_2015}. The equation for forward KL divergence is given by:
\begin{equation}
\label{eq:fwd_kld}
    D_{KL}(P \| Q) = \sum_{x} P(x) \log \left( \frac{P(x)}{Q(x)} \right)
\end{equation}
From Equation \ref{eq:fwd_kld} we see that forward KL divergence is minimized in a given region when the value of $P$ is low and $Q$ is high. This results in students exhibiting \textit{mean-seeking} behavior, with diverse but potentially nonsensical responses \citep{gu_minillm_2024}.

\paragraph{Reverse KL Divergence} Recent work has explored using reverse KL divergence as an alternative to the forward variant in generative models \citep{gu_minillm_2024, huszar_how_2015, nowozin_f-gan_2016, chen_symmetric_2017, lee_self-knowledge_2022}. The equation for reverse KL divergence is given by:
\begin{equation}
\label{eq:rev_kld}
    D_{KL}(Q \| P) = \sum_{x} Q(x) \log \left( \frac{Q(x)}{P(x)} \right)
\end{equation}
In contrast to forward, reverse KL divergence is minimized in a region when the value of the approximating distribution $Q$ is low and the true distribution $P$ is high. In practice, reverse KL divergence produces \textit{mode-seeking} students which may lack diversity, but can be highly specific and accurate \citep{gu_minillm_2024}.

\subsubsection{Jensen-Shannon Divergence}
\label{sss:jsd}
Though closely related to KL divergence, Jensen-Shannon (JS) divergence has only rarely been applied to KD \citep{wen_f-divergence_2023}. Its equation is given by:
\begin{equation}
\label{eq:jsd}
    D_{JS}(P \| Q) = \frac{1}{2} D_{KL}(P \| M) + \frac{1}{2} D_{KL}(Q \| M)
\end{equation}
where
\begin{equation*}
M = \frac{P+Q}{2}.
\end{equation*}

JS divergence can be thought of as an average of forward and reverse KL divergence, producing students which can capture more diversity than reverse KL divergence and produce more faithful responses than forward KL divergence.

\subsubsection{Generalizing Divergence} 
Using the mixing intuition from Section \ref{sss:jsd}, divergence can be generalized using parameter $\beta$ to control the weighting between forward and reverse KL divergence:
\begin{equation}
\label{eq:beta-jsd}
    D_{JS}^{\beta}(P \| Q) = \beta \cdot D_{KL}(P \| M_{\beta}) + (1-\beta) \cdot D_{KL}(Q \| M_{\beta})
\end{equation}
where,
\begin{equation*}
\label{eq:mixed_dist}
M_{\beta} = \beta P + (1-\beta)Q.
\end{equation*}
This facilitates tuning $\beta$ to balance requirements for response diversity and faithfulness.

\subsection{Mechanistic Interpretability}
The field of mechanistic interpretability works to build a human-interpretable understanding of LLMs, and has been able to develop some intuition as to how different transformer components contribute to model processing. The residual stream represents the accumulation of model knowledge, and can be thought of as a checkpoint or "best guess" at any point in a transformer. Attention is the only mechanism by which information mixes across positions in the residual stream, and can be understood as a syntactically-informed and context-aware generalized convolution. The multilayer perceptron (MLP), though challenging to cleanly interpret due to superposition \citep{elhage_toy_2022, ameisen_circuit_2025, templeton_scaling_2024} can be thought of as the information processing core of the transformer architecture, operating independently on each position in the residual stream \citep{elhage_mathematical_2021}.

\section{Related Work}
\subsection{Distillation Objective Functions}
Distillation objective functions in white-box KD have taken many different forms since the original forward KL divergence implementation from \citet{hinton_distilling_2015}. Simpler versions of LLM distillation use a multi-objective loss function which combines the soft targets with the true language modeling loss \citep{xu_survey_2024}. The DistilBERT model expanded upon this to also add a cosine embedding loss to encourage representational alignment between the student and teacher, forming a triple-objective \citep{sanh_distilbert_2020}. The f-Distill framework was designed with modularity to allow comparison between forward KL divergence, reverse KL divergence, JS divergence, and total variance distance \citep{wen_f-divergence_2023}. MiniLLM is a more recent example which used reverse KL divergence as the objective in sequence-based KD for on-policy reinforcement learning in modern multi-billion parameter LLMs \citep{gu_minillm_2024}.

\subsection{Distilling Internal Activations}
The success of distilling internal representations is well-documented, including in deep convolutional networks \citep{romero_fitnets_2015} and BERT models \citep{sun_patient_2019, jiao_tinybert_2020, hou_dynabert_2020, liang_homodistil_2023, zuo_moebert_2022}. Two recent works which distilled activations in each transformer block are TinyBERT \citep{jiao_tinybert_2020} and TED \citep{liang_less_2023}. 

TinyBERT distilled the QK attention matrices of BERT's multi-head attention and the final block outputs into each block of the student model, in addition to the embeddings and the regular CE language modeling loss. While this is an efficient approach to distilling syntactic knowledge, it is not very generalizable. QK distillation is only viable when the student has the same number of attention heads as the teacher.

TED was designed to build upon layer-wise distillation by training filters to remove redundant information from the residual stream. However, the concept of redundant information in transformers was explored only at the scale of a single layer or single neuron in encoder-only models \citep{durrani_analyzing_2020, dalvi_analyzing_2020, antverg_pitfalls_2022}. Recent work studying cross-layer superposition with crosscoders and transcoders in modern decoder-only models has demonstrated that such simple notions of redundancy may fail to capture more complex information storage and processing in LLMs \citep{ameisen_circuit_2025}. Although TED demonstrated success on GPT-2 Small (124M parameters) \citep{radford_language_2019}, it remains to be seen how it scales to modern 1B+ models and benchmarks.

\section{Distillation as a Parametrized System}
\label{s:div_param_sys}
In this work, we take a first-principles approach to developing a new SOTA methodology for knowledge distillation in LLMs. We identify representations- both internal activations and output distributions- as the core units of knowledge which we aim to distill. It naturally follows that we also focus on the divergence and similarity paradigm of KD \citep{xu_survey_2024}, as it is best suited to make use of model internals.

Some recent works such as MiniLLM \citep{gu_minillm_2024} and the f-DISTILL framework \citep{wen_f-divergence_2023} explore knowledge distillation at the sequence level. While we recognize the demonstrated utility of sequence-level KD, sequence generation is inefficient for full and continual pretraining in modern settings, and thus we focus on token-level KD. 

We identify four parameters of a general KD method: student initialization, activation projection, knowledge gap measure, and objective function. A summary of the explorations we conducted for each of these variables is included in Appendix \ref{a:exploration}.

\subsection{Student Initialization}
\label{ss:student_init}
Much of the prior work in white-box KD initializes students as \textit{shallow} versions of the teacher model, with a reduced layer count, but otherwise identical architecture \citep{xu_survey_2024, jiao_tinybert_2020, liang_less_2023, wen_f-divergence_2023, sanh_distilbert_2020}. This makes intuitive sense, as the student model is maximally compatible with the teacher, allowing for distillation of attention heads and removing the need to a projection between embedding dimension. However, these works fully pretrain their students to recover from the architectural disruption- an infeasible process in many modern settings. Our exploration found that shallow students performed very poorly with the given training budgets, and thus we do not include results for shallow initialization in Section \ref{s:results}. Details on the poor performance of shallow students can be found in Appendix \ref{ss:stage1}.

The alternative is a pretrained student, which sacrifices full compatibility for a superior initialization point. The activations of pretrained students are often not in the same embedding space as the teacher, making a projection necessary to measure the knowledge gap. Attention head activations are also incompatible for pretrained student distillation\footnote{See Appendix \ref{a:incompat_attn_heads} for further discussion}. Unlike shallow students, the initialization point for pretrained students represents a highly capable model, so we include comparisons against the Baseline and finetuned (FT) students in Section \ref{s:results}.

\subsection{Activation Projection}
In the case where the student and teacher embedding dimensions are different, distilling internal activations requires projecting activations into a shared space to compute loss during training. In prior works, this was done by inserting a trainable matrix to perform a linear projection of the student activations into the teacher's embedding space. We refer to this as an \textit{upward-online} projection, as the output of the projection is in a higher embedding space than its input, and the projection matrix is learned "online" during training. We explored a variety of projectors and found this formulation to be the best. Details regarding the different projector architectures we experimented with can be found in Appendix \ref{a:projection}, and our projection exploration is described in Appendices \ref{ss:stage1} and \ref{ss:stage2}. 

\subsection{Knowledge Gap Measure}
Knowledge distillation can be set up as a problem of distribution-matching between student and teacher outputs. This intuitively leads to the use of forward KL divergence as the loss measure, as it sets the teacher outputs as the target distribution, and a loss value of 0 indicates a perfect match between the teacher and student. Recent works such as MiniLLM \citep{gu_minillm_2024} and f-DISTILL \citep{wen_f-divergence_2023} have explored using different measures, including reverse KL divergence, citing the disadvantages of mean-seeking behavior in generative tasks. 

Instead of using a strict forward or reverse setting, we explored a range of target distribution mixing ratios to empirically determine the best divergence measure. Our exploration of the parametrized divergence from Equation \ref{eq:beta-jsd} is detailed in Appendix \ref{a:div_search}. 

\subsection{Objective Function}
A KD objective function should contain all of the important and useful signals from the teacher intended to be replicated by the student. The most obvious signal is the output distribution, since it indicates the output as well as a summary of the model's comprehension, as indicated by other high-probability predictions. Since it can be difficult to train a student from the output alone, previous works have included internal activations- such as residual outputs- at different points throughout the model. These can intuitively be thought of as checkpoints in model processing. Motivated by the findings of mechanistic interpretability, we explored new formulations of the objective function. Details can be found in Appendix \ref{a:objective}.

\section{Methodology}
KDFP utilizes a pretrained student initialization. Activations are projected with an upward-online projector. It uses a divergence parameter of 0.25, corresponding to a 75\% teacher, 25\% student mixing ratio for the target distribution. Given $N$ blocks in a student model, KDFP maps $N$ linearly-spaced blocks from the teacher model to the student. KDFP distills the outputs of the attention, MLP, and residual stream at every block in the student, as well as the output distribution and ground truth (i.e. cross-entropy) loss. All objective components are weighted equally ($\alpha = \frac{1}{6}$).  

\subsection{Comparisons to Other Distillation Methods}
We compare the KDFP methodology against the traditional KD formulation from \citet{hinton_distilling_2015}, layer-wise distillation (LWD)\footnote{Unlike \citet{liang_less_2023} we do not train a projector for each layer, making our LWD variant less expressive than the one previously beaten by TED in \citet{liang_less_2023}}, and TED \citep{liang_less_2023} in the context of the four parameters outlined in Section \ref{s:div_param_sys}.

We use pretrained initialization for all students. Traditional KD, LWD, and TED use forward KL divergence as the measure of the knowledge gap. Traditional KD does not distill internal activations, so no projection is necessary. TED performs a projection in its task-aware filters to produce outputs in the teacher's embedding space (i.e. an upward projection). LWD follows an identical approach to KDFP, using a single upward-online projector, except that LWD only distills the residual. 

All of these methods utilize cross-entropy (CE) loss on the labels and a divergence loss. Nothing else is added for traditional KD. LWD includes the mean squared error (MSE) loss on the residual activations. TED\footnote{We use the 2-layer MLP filters for the best zero-shot performance \citep{liang_less_2023}} follows a multi-stage training process. In the warmup stage, filters are trained for the teacher and student using CE. In the distillation phase, TED replaces the raw residual outputs of LWD with the outputs of the task-aware filters. A summary of the comparison of distillation methods illustrated in Table \ref{tab:method-comparison}.

\begin{table}[tbp]
\centering
\small
\begin{tabular}{lllll}
\toprule
\textbf{Feature} & \textbf{Traditional KD} & \textbf{LWD} & \textbf{TED} & \textbf{KDFP (Ours)} \\
\midrule
\textbf{Internal Distill.} & No & Yes & Yes & Yes \\
\textbf{Projection} & None & Up-Online & Up (Task Filters) & Up-Online \\
\textbf{Projector Count} & N/A & Single & Per-layer & Single \\
\textbf{Objectives} & CE, Div & CE, Div, MSE & CE, Div, MSE & CE, Div, MSE \\
\textbf{Divergence} & Forward KL & Forward KL & Forward KL & Param. Div. ($\beta=0.25$) \\
\bottomrule
\end{tabular}
\caption{Comparison of distillation methodologies}
\label{tab:method-comparison}
\end{table}

\section{Experiments}
We evaluate KDFP in the continual pretraining setting, with the standard LLM task of causal language modeling. In Section \ref{ss:main_results}, we use teachers which have not been finetuned on the dataset to reflect the setting where budget or resource constraints may not allow for teacher finetuning. Analysis of the impact of teacher finetuning is presented in Section \ref{ss:ft_teacher}. As discussed in Section \ref{ss:student_init}, Baseline and FT represent clear thresholds that all KD methods are expected to surpass. 

\paragraph{Models}
We evaluate KDFP using Pythia 2.8B and 1.4B\footnote{Common modern student sizes are in the 1B - 7B range. Model selection was influenced by resource constraints and the ability to scale within the same model family. See Section \ref{ss:limits} for further discussion.} \citep{biderman_pythia_2023} for the teacher and student models, respectively. All Pythia models are the deduped variants.

\paragraph{Datasets}
We conduct all of our training using data from  FineWeb Edu \citep{penedo_fineweb_2024}. It is a recent, high-quality dataset designed as an LLM pretraining corpus. \added{Using a subset of a new, high-quality dataset in a continual pretraining scenario is an efficient approach to producing a more capable model with a far lower cost than fully pretraining a new model with the entire new dataset.}

We train both teachers and students on 100M tokens with a sequence length of 2048- the full context for the Pythia models. TED requires a warmup stage for the teacher and student filters. We train the filters for 25M tokens to replicate the 1:4 filter-to-distillation training data ratio from \citet{liang_less_2023}.

\paragraph{Evaluation}
We evaluate each student across 9 common modern benchmarks \citep{xu_survey_2024, zhu_survey_2024, team_gemma_2025, grattafiori_llama_2024}, all zero-shot: accuracy on BoolQ \citep{clark_boolq_2019}, MMLU \citep{hendrycks_measuring_2021}, and WinoGrande (WG) \citep{sakaguchi_winogrande_2019}; normalized accuracy on ARC-Easy and ARC-Challenge \citep{clark_think_2018}, PIQA \citep{bisk_piqa_2019}, and SciQ \citep{johannes_welbl_crowdsourcing_2017}; F1 score on DROP \citep{dua_drop_2019}; and exact match on TriviaQA (TQA) \citep{joshi_triviaqa_2017}. We present ARC as the sum of ARC-Challenge and ARC-Easy due to space constraints. We ran our benchmarks using EleutherAI's LM Evaluation Harness \citep{gao_language_2024}.

\section{Results}
\label{s:results}

\subsection{Main Results}
\label{ss:main_results}
From Table \ref{tab:main_results} it is clear that TED does not scale well to modern models and evaluations, degrading the performance of the Baseline student. By training for an additional 0.03\% of the pretraining token budget, KDFP boosts average performance across all 9 benchmarks by nearly 5\% relative to TED, and 2.9\% relative to the Baseline. The results from KDFP here represent the \textit{peak average score} observed in all of our experiments (0.4358), indicating a clear advantage across a diversity of tasks.

\begin{table}[tbp]
    \centering
    \small
    \begin{tabular}{lccccccccr}
        \textbf{Method} & \textbf{ARC} & \textbf{BoolQ} & \textbf{DROP} & \textbf{MMLU} & \textbf{PIQA} & \textbf{SciQ} & \textbf{TQA} & \textbf{WG} & \textbf{Avg.}\\
        \toprule
        Baseline & 0.8469 & 0.5771 & 0.0280 & 0.2361 & 0.7187 & 0.773 & 0.0760 & 0.5572 & 0.4236 \\
        FT       & 0.8170 & 0.6110 & 0.0425 & \textbf{0.2398} & 0.7144 & 0.785 & 0.0244 & \textbf{0.5706} & 0.4227 \\
        KD       & 0.8326 & 0.5936 & 0.0367 & 0.2319 & \textbf{0.7231} & 0.785 & 0.0388 & 0.5675 & 0.4232 \\
        LWD      & 0.8432 & 0.6235 & \textbf{0.0463} & 0.2316 & 0.7203 & 0.785 & 0.0636 & 0.5485 & 0.4291 \\
        TED      & 0.8063 & 0.6125 & 0.0247 & 0.2294 & 0.7040 & 0.765 & 0.0378 & 0.5588 & 0.4154 \\
        \midrule
        KDFP     & \textbf{0.8542} & \textbf{0.6291} & \textbf{0.0463} & 0.2395 & 0.7182 & \textbf{0.798} & \textbf{0.0762} & 0.5612 & \textbf{0.4358} \\
        \bottomrule
\end{tabular}
    \caption{Main results. 100M tokens of data and 2.8B parameter teacher without finetuning.}
    \label{tab:main_results}
\end{table}

\subsection{Teacher Finetuning}
\label{ss:ft_teacher}
In Table \ref{tab:ft_teacher} we evaluate KDFP with a teacher finetuned on the dataset (FineWeb Edu) prior to distillation. 
LWD saw notable gains, though still falls behind the peak average score by 1.3\%.
KDFP is again the top performer, though we find that the peak average score here dropped by $\sim$1\%, indicating that finetuned teachers do not necessarily benefit students.
We hypothesize that the distribution drift from finetuning may make it more challenging for the student to learn the teacher's representations, with KDFP suffering relative to LWD due to its increased number of distilled activations.

\subsection{Teacher Scaling}
\label{ss:big_teacher}
In Table \ref{tab:big_teacher} we analyze how scaling up the teacher to Pythia 6.9B affects student performance. Notably, the only architectural difference between Pythia 2.8B and 6.9B is the embedding dimension\footnote{Attention head dimension is technically larger, but only because the embeddings are larger.}. 
Similar to finetuning the teacher, using larger teachers largely did not help students. KDFP again performed best with TED showing a very small improvement, but performance dropped of 1.3\% relative to the peak average score.While the embedding dimension of Pythia 2.8B is only 25\% larger than the student, 6.9B is fully double (4096 vs. 2048), making it difficult for students to learn from teacher representations.

\subsection{Data Scaling}
\label{ss:5x_data}
In Table \ref{tab:5x_data} we evaluate KDFP with a 5x increase in data (500M tokens). The filter warmup stages for TED were increased to 125M tokens to maintain the same data ratio. 
We again see a peak performance drop, falling 2.1\% behind the peak average score. Though LWD performs best and FT improves, KDFP in Section \ref{ss:main_results} outperforms all of these with an 80\% reduction in training cost.

\begin{table}[tbp]
    \centering
    \small
        \begin{subtable}{0.32\textwidth}
        \centering
        \begin{tabular}{lr}
             \textbf{Method} & \textbf{Avg. Score}  \\
             \toprule
             KD & 0.4232 \\
             LWD & \textit{0.4303} \\
             TED & 0.4140 \\
             KDFP & \textbf{0.4316} \\
             \bottomrule
        \end{tabular}
        \caption{Finetuned 2.8B Teacher}
        \label{tab:ft_teacher}
    \end{subtable}
    \hfill
    \begin{subtable}{0.32\textwidth}
        \centering
        \begin{tabular}{lr}
             \textbf{Method} & \textbf{Avg. Score}  \\
             \toprule
             KD & 0.4232 \\
             LWD & 0.4284 \\
             TED & \textit{0.4165} \\
             KDFP & \textbf{0.4303} \\
             \bottomrule
        \end{tabular}
        \caption{6.9B Teacher}
        \label{tab:big_teacher}
    \end{subtable}
    \hfill
    \begin{subtable}{0.32\textwidth}
        \centering
        \begin{tabular}{lr}
             \textbf{Method} & \textbf{Avg. Score}  \\
             \toprule
             FT & \textit{0.4263} \\
             KD & 0.4207 \\
             LWD & \textbf{0.4266} \\
             TED & 0.4154 \\
             KDFP &  0.4264 \\
             \bottomrule
        \end{tabular}
        \caption{5x Data}
        \label{tab:5x_data}
    \end{subtable}
    \caption{Analyses of KDFP. 
    Table \ref{tab:ft_teacher} shows finetuning the teacher prior to distillation.
    Table \ref{tab:big_teacher} shows scaling the teacher from 2.8B to 6.9B parameters. 
    Table \ref{tab:5x_data} shows scaling the training data to 500M tokens. 
    Italics indicate an improvement over the Main Results (Table \ref{tab:main_results}). 
    Full results for these experiments are provided in Appendix \ref{a:extended_analysis}.}
    \hfill
\end{table}

\section{Discussion}

\subsection{Training Efficiency}
\label{ss:efficiency}
Once of the strengths of KDFP is its efficiency.
KDFP significantly reduces the number of ephemeral parameters trained as part of the distillation relative to TED. With Pythia 2.8B and 1.4B only 5.2M temporary parameters are trained (i.e. the projector), while nearly 600M temporary parameters are trained by TED (i.e. the filters). Even with such relatively small models, KDFP reduces training requirements by over half a billion parameters. 
Although KDFP utilizes more activations than other methods, these activations are already computed as part of the forward pass. Thus, the only increase in compute comes from storing and computing losses on activations.

KDFP makes better use of information which is already available from small, unmodified teachers and doesn't require long, expensive training schedules. KDFP is not only successful with an efficient setup, our results indicate that this is \textit{preferred}. Given the option between more and less efficient options, KDFP appears to thrive with efficiency. We leave it to future work to test the limits of efficiency in KDFP.

\subsection{Explaining TED Performance}
\label{ss:explaining_ted}
Recent work in mechanistic interpretability has shown that LLMs likely store their knowledge across layers \citep{elhage_toy_2022, lindsey_biology_2025, lindsey_sparse_2024, ameisen_circuit_2025}. We hypothesize that TED's attempt to capture and filter out redundant information may actually be disrupting the information processing. Additionally, training with filters along the residual stream and then removing them for inference likely also introduces disruption. All of the experiments in Section \ref{s:results} indicate that TED degrades performance below the Baseline pretrained student. This serves as an indicator that our hypothesis may be correct, and future work may further investigate.

\subsection{Limitations}
\label{ss:limits}

\paragraph{Continual vs. Full Pretraining}
Due to resource constraints, we are unable to test KDFP in the full pretraining setting and thus we only evaluate it in continual pretraining. Even though the student models we train are considered small at less than 2B parameters, it is still infeasible for us to pretrain models of this size with the resources at our disposal. We leave the evaluation of KDFP in the full pretraining setting to future work by those with sufficient computational resources.

\paragraph{Implementation Challenges}
Our implementation of KDFP utilizes the TransformerLens library \citep{nanda_transformerlens_2022}, which standardizes internal activation extraction but does not fully support multi-GPU setups. This limited our experiments to models which could be trained on a single GPU. While small, our 1.4B student is comparable in parameter count to the 1.8B parameter Gemini Nano-1\footnote{The performance of our Pythia 1.4B students is not directly comparable to Nano-1, as Nano-1 is 4-bit quantized.} \citep{team_gemini_2025}, a model commercially deployed on edge devices. For those who require multi-GPU support or models not currently supported by TransformerLens, activation extraction with HuggingFace's \texttt{transformers} library \citep{wolf_huggingfaces_2019} is still feasible, but not standardized.

\section{Conclusion}
We introduced KDFP, a new method for distilling LLMs, and demonstrated its superior performance and efficiency relative to comparable methods. KDFP demonstrates that more fully utilizing the information available in transformers produces more capable students, and that a target distribution ratio of 75\% teacher, 25\% student outperforms existing methods. We found that less data and teacher--student alignment may be beneficial for distillation, as KDFP requires only 100M tokens of training with a small, unmodified, teacher for peak performance. KDFP is an example of how better understanding the science of LLMs can lead to practical improvements.

\newpage

\bibliography{references}

@misc{ameisen_circuit_2025,
	title = {Circuit {Tracing}: {Revealing} {Computational} {Graphs} in {Language} {Models}},
	shorttitle = {Circuit {Tracing}},
	url = {https://transformer-circuits.pub/2025/attribution-graphs/methods.html},
	language = {en},
	urldate = {2025-12-01},
	journal = {Transformer Circuits},
	author = {Ameisen*, Authors Emmanuel and Lindsey*, Jack and Pearce*, Adam and Gurnee*, Wes and Turner*, Nicholas L. and Chen*, Brian and Citro*, Craig and Abrahams, David and Carter, Shan and Hosmer, Basil and Marcus, Jonathan and Sklar, Michael and Templeton, Adly and Bricken, Trenton and McDougall, Callum and Cunningham, Hoagy and Henighan, Thomas and Jermyn, Adam and Jones, Andy and Persic, Andrew and Qi, Zhenyi and Thompson, T. Ben and Zimmerman, Sam and Rivoire, Kelley and Conerly, Thomas and Olah, Chris and March 27, Joshua Batson*‡ Affiliations Anthropic Published},
	month = mar,
	year = {2025},
}

@article{radford_language_2019,
	title = {Language {Models} are {Unsupervised} {Multitask} {Learners}},
	language = {en},
	author = {Radford, Alec and Wu, Jeffrey and Child, Rewon and Luan, David and Amodei, Dario and Sutskever, Ilya},
	year = {2019},
}

@misc{zhu_survey_2024,
	title = {A {Survey} on {Model} {Compression} for {Large} {Language} {Models}},
	url = {https://arxiv.org/abs/2308.07633},
	author = {Zhu, Xunyu and Li, Jian and Liu, Yong and Ma, Can and Wang, Weiping},
	year = {2024},
}

@inproceedings{clark_boolq_2019,
	title = {{BoolQ}: {Exploring} the {Surprising} {Difficulty} of {Natural} {Yes}/{No} {Questions}},
	booktitle = {{NAACL}},
	author = {Clark, Christopher and Lee, Kenton and Chang, Ming-Wei and Kwiatkowski, Tom and Collins, Michael and Toutanova, Kristina},
	year = {2019},
}

@article{templeton_scaling_2024,
	title = {Scaling {Monosemanticity}: {Extracting} {Interpretable} {Features} from {Claude} 3 {Sonnet}},
	url = {https://transformer-circuits.pub/2024/scaling-monosemanticity/index.html},
	journal = {Transformer Circuits Thread},
	author = {Templeton, Adly and Conerly, Tom and Marcus, Jonathan and Lindsey, Jack and Bricken, Trenton and Chen, Brian and Pearce, Adam and Citro, Craig and Ameisen, Emmanuel and Jones, Andy and Cunningham, Hoagy and Turner, Nicholas L and McDougall, Callum and MacDiarmid, Monte and Freeman, C. Daniel and Sumers, Theodore R. and Rees, Edward and Batson, Joshua and Jermyn, Adam and Carter, Shan and Olah, Chris and Henighan, Tom},
	year = {2024},
}

@inproceedings{liang_less_2023,
	address = {Honolulu, Hawaii, USA},
	series = {{ICML}'23},
	title = {Less is more: task-aware layer-wise distillation for language model compression},
	volume = {202},
	shorttitle = {Less is more},
	urldate = {2025-10-13},
	booktitle = {Proceedings of the 40th {International} {Conference} on {Machine} {Learning}},
	publisher = {JMLR.org},
	author = {Liang, Chen and Zuo, Simiao and Zhang, Qingru and He, Pengcheng and Chen, Weizhu and Zhao, Tuo},
	month = jul,
	year = {2023},
	pages = {20852--20867},
}

@inproceedings{johannes_welbl_crowdsourcing_2017,
	title = {Crowdsourcing {Multiple} {Choice} {Science} {Questions}},
	booktitle = {{arXiv}:1707.06209v1},
	author = {Johannes Welbl, Nelson F. Liu, Matt Gardner},
	year = {2017},
}

@misc{gao_language_2024,
	title = {The {Language} {Model} {Evaluation} {Harness}},
	url = {https://zenodo.org/records/12608602},
	doi = {10.5281/zenodo.12608602},
	publisher = {Zenodo},
	author = {Gao, Leo and Tow, Jonathan and Abbasi, Baber and Biderman, Stella and Black, Sid and DiPofi, Anthony and Foster, Charles and Golding, Laurence and Hsu, Jeffrey and Le Noac'h, Alain and Li, Haonan and McDonell, Kyle and Muennighoff, Niklas and Ociepa, Chris and Phang, Jason and Reynolds, Laria and Schoelkopf, Hailey and Skowron, Aviya and Sutawika, Lintang and Tang, Eric and Thite, Anish and Wang, Ben and Wang, Kevin and Zou, Andy},
	month = jul,
	year = {2024},
	note = {Version Number: v0.4.3},
}

@article{elhage_mathematical_2021,
	title = {A mathematical framework for transformer circuits},
	volume = {1},
	number = {1},
	journal = {Transformer Circuits Thread},
	author = {Elhage, Nelson and Nanda, Neel and Olsson, Catherine and Henighan, Tom and Joseph, Nicholas and Mann, Ben and Askell, Amanda and Bai, Yuntao and Chen, Anna and Conerly, Tom and {others}},
	year = {2021},
	pages = {12},
}

@inproceedings{biderman_pythia_2023,
	title = {Pythia: {A} suite for analyzing large language models across training and scaling},
	booktitle = {International {Conference} on {Machine} {Learning}},
	publisher = {PMLR},
	author = {Biderman, Stella and Schoelkopf, Hailey and Anthony, Quentin Gregory and Bradley, Herbie and O’Brien, Kyle and Hallahan, Eric and Khan, Mohammad Aflah and Purohit, Shivanshu and Prashanth, USVSN Sai and Raff, Edward and {others}},
	year = {2023},
	pages = {2397--2430},
}

@misc{team_gemini_2025,
	title = {Gemini: {A} {Family} of {Highly} {Capable} {Multimodal} {Models}},
	shorttitle = {Gemini},
	url = {http://arxiv.org/abs/2312.11805},
	doi = {10.48550/arXiv.2312.11805},
	urldate = {2026-03-26},
	publisher = {arXiv},
	author = {Team, Gemini and Anil, Rohan and Borgeaud, Sebastian and Alayrac, Jean-Baptiste and Yu, Jiahui and Soricut, Radu and Schalkwyk, Johan and Dai, Andrew M. and Hauth, Anja and Millican, Katie and Silver, David and Johnson, Melvin and Antonoglou, Ioannis and Schrittwieser, Julian and Glaese, Amelia and Chen, Jilin and Pitler, Emily and Lillicrap, Timothy and Lazaridou, Angeliki and Firat, Orhan and Molloy, James and Isard, Michael and Barham, Paul R. and Hennigan, Tom and Lee, Benjamin and Viola, Fabio and Reynolds, Malcolm and Xu, Yuanzhong and Doherty, Ryan and Collins, Eli and Meyer, Clemens and Rutherford, Eliza and Moreira, Erica and Ayoub, Kareem and Goel, Megha and Krawczyk, Jack and Du, Cosmo and Chi, Ed and Cheng, Heng-Tze and Ni, Eric and Shah, Purvi and Kane, Patrick and Chan, Betty and Faruqui, Manaal and Severyn, Aliaksei and Lin, Hanzhao and Li, YaGuang and Cheng, Yong and Ittycheriah, Abe and Mahdieh, Mahdis and Chen, Mia and Sun, Pei and Tran, Dustin and Bagri, Sumit and Lakshminarayanan, Balaji and Liu, Jeremiah and Orban, Andras and Güra, Fabian and Zhou, Hao and Song, Xinying and Boffy, Aurelien and Ganapathy, Harish and Zheng, Steven and Choe, HyunJeong and Weisz, Ágoston and Zhu, Tao and Lu, Yifeng and Gopal, Siddharth and Kahn, Jarrod and Kula, Maciej and Pitman, Jeff and Shah, Rushin and Taropa, Emanuel and Merey, Majd Al and Baeuml, Martin and Chen, Zhifeng and Shafey, Laurent El and Zhang, Yujing and Sercinoglu, Olcan and Tucker, George and Piqueras, Enrique and Krikun, Maxim and Barr, Iain and Savinov, Nikolay and Danihelka, Ivo and Roelofs, Becca and White, Anaïs and Andreassen, Anders and Glehn, Tamara von and Yagati, Lakshman and Kazemi, Mehran and Gonzalez, Lucas and Khalman, Misha and Sygnowski, Jakub and Frechette, Alexandre and Smith, Charlotte and Culp, Laura and Proleev, Lev and Luan, Yi and Chen, Xi and Lottes, James and Schucher, Nathan and Lebron, Federico and Rrustemi, Alban and Clay, Natalie and Crone, Phil and Kocisky, Tomas and Zhao, Jeffrey and Perz, Bartek and Yu, Dian and Howard, Heidi and Bloniarz, Adam and Rae, Jack W. and Lu, Han and Sifre, Laurent and Maggioni, Marcello and Alcober, Fred and Garrette, Dan and Barnes, Megan and Thakoor, Shantanu and Austin, Jacob and Barth-Maron, Gabriel and Wong, William and Joshi, Rishabh and Chaabouni, Rahma and Fatiha, Deeni and Ahuja, Arun and Tomar, Gaurav Singh and Senter, Evan and Chadwick, Martin and Kornakov, Ilya and Attaluri, Nithya and Iturrate, Iñaki and Liu, Ruibo and Li, Yunxuan and Cogan, Sarah and Chen, Jeremy and Jia, Chao and Gu, Chenjie and Zhang, Qiao and Grimstad, Jordan and Hartman, Ale Jakse and Garcia, Xavier and Pillai, Thanumalayan Sankaranarayana and Devlin, Jacob and Laskin, Michael and Casas, Diego de Las and Valter, Dasha and Tao, Connie and Blanco, Lorenzo and Badia, Adrià Puigdomènech and Reitter, David and Chen, Mianna and Brennan, Jenny and Rivera, Clara and Brin, Sergey and Iqbal, Shariq and Surita, Gabriela and Labanowski, Jane and Rao, Abhi and Winkler, Stephanie and Parisotto, Emilio and Gu, Yiming and Olszewska, Kate and Addanki, Ravi and Miech, Antoine and Louis, Annie and Teplyashin, Denis and Brown, Geoff and Catt, Elliot and Balaguer, Jan and Xiang, Jackie and Wang, Pidong and Ashwood, Zoe and Briukhov, Anton and Webson, Albert and Ganapathy, Sanjay and Sanghavi, Smit and Kannan, Ajay and Chang, Ming-Wei and Stjerngren, Axel and Djolonga, Josip and Sun, Yuting and Bapna, Ankur and Aitchison, Matthew and Pejman, Pedram and Michalewski, Henryk and Yu, Tianhe and Wang, Cindy and Love, Juliette and Ahn, Junwhan and Bloxwich, Dawn and Han, Kehang and Humphreys, Peter and Sellam, Thibault and Bradbury, James and Godbole, Varun and Samangooei, Sina and Damoc, Bogdan and Kaskasoli, Alex and Arnold, Sébastien M. R. and Vasudevan, Vijay and Agrawal, Shubham and Riesa, Jason and Lepikhin, Dmitry and Tanburn, Richard and Srinivasan, Srivatsan and Lim, Hyeontaek and Hodkinson, Sarah and Shyam, Pranav and Ferret, Johan and Hand, Steven and Garg, Ankush and Paine, Tom Le and Li, Jian and Li, Yujia and Giang, Minh and Neitz, Alexander and Abbas, Zaheer and York, Sarah and Reid, Machel and Cole, Elizabeth and Chowdhery, Aakanksha and Das, Dipanjan and Rogozińska, Dominika and Nikolaev, Vitaliy and Sprechmann, Pablo and Nado, Zachary and Zilka, Lukas and Prost, Flavien and He, Luheng and Monteiro, Marianne and Mishra, Gaurav and Welty, Chris and Newlan, Josh and Jia, Dawei and Allamanis, Miltiadis and Hu, Clara Huiyi and Liedekerke, Raoul de and Gilmer, Justin and Saroufim, Carl and Rijhwani, Shruti and Hou, Shaobo and Shrivastava, Disha and Baddepudi, Anirudh and Goldin, Alex and Ozturel, Adnan and Cassirer, Albin and Xu, Yunhan and Sohn, Daniel and Sachan, Devendra and Amplayo, Reinald Kim and Swanson, Craig and Petrova, Dessie and Narayan, Shashi and Guez, Arthur and Brahma, Siddhartha and Landon, Jessica and Patel, Miteyan and Zhao, Ruizhe and Villela, Kevin and Wang, Luyu and Jia, Wenhao and Rahtz, Matthew and Giménez, Mai and Yeung, Legg and Keeling, James and Georgiev, Petko and Mincu, Diana and Wu, Boxi and Haykal, Salem and Saputro, Rachel and Vodrahalli, Kiran and Qin, James and Cankara, Zeynep and Sharma, Abhanshu and Fernando, Nick and Hawkins, Will and Neyshabur, Behnam and Kim, Solomon and Hutter, Adrian and Agrawal, Priyanka and Castro-Ros, Alex and Driessche, George van den and Wang, Tao and Yang, Fan and Chang, Shuo-yiin and Komarek, Paul and McIlroy, Ross and Lučić, Mario and Zhang, Guodong and Farhan, Wael and Sharman, Michael and Natsev, Paul and Michel, Paul and Bansal, Yamini and Qiao, Siyuan and Cao, Kris and Shakeri, Siamak and Butterfield, Christina and Chung, Justin and Rubenstein, Paul Kishan and Agrawal, Shivani and Mensch, Arthur and Soparkar, Kedar and Lenc, Karel and Chung, Timothy and Pope, Aedan and Maggiore, Loren and Kay, Jackie and Jhakra, Priya and Wang, Shibo and Maynez, Joshua and Phuong, Mary and Tobin, Taylor and Tacchetti, Andrea and Trebacz, Maja and Robinson, Kevin and Katariya, Yash and Riedel, Sebastian and Bailey, Paige and Xiao, Kefan and Ghelani, Nimesh and Aroyo, Lora and Slone, Ambrose and Houlsby, Neil and Xiong, Xuehan and Yang, Zhen and Gribovskaya, Elena and Adler, Jonas and Wirth, Mateo and Lee, Lisa and Li, Music and Kagohara, Thais and Pavagadhi, Jay and Bridgers, Sophie and Bortsova, Anna and Ghemawat, Sanjay and Ahmed, Zafarali and Liu, Tianqi and Powell, Richard and Bolina, Vijay and Iinuma, Mariko and Zablotskaia, Polina and Besley, James and Chung, Da-Woon and Dozat, Timothy and Comanescu, Ramona and Si, Xiance and Greer, Jeremy and Su, Guolong and Polacek, Martin and Kaufman, Raphaël Lopez and Tokumine, Simon and Hu, Hexiang and Buchatskaya, Elena and Miao, Yingjie and Elhawaty, Mohamed and Siddhant, Aditya and Tomasev, Nenad and Xing, Jinwei and Greer, Christina and Miller, Helen and Ashraf, Shereen and Roy, Aurko and Zhang, Zizhao and Ma, Ada and Filos, Angelos and Besta, Milos and Blevins, Rory and Klimenko, Ted and Yeh, Chih-Kuan and Changpinyo, Soravit and Mu, Jiaqi and Chang, Oscar and Pajarskas, Mantas and Muir, Carrie and Cohen, Vered and Lan, Charline Le and Haridasan, Krishna and Marathe, Amit and Hansen, Steven and Douglas, Sholto and Samuel, Rajkumar and Wang, Mingqiu and Austin, Sophia and Lan, Chang and Jiang, Jiepu and Chiu, Justin and Lorenzo, Jaime Alonso and Sjösund, Lars Lowe and Cevey, Sébastien and Gleicher, Zach and Avrahami, Thi and Boral, Anudhyan and Srinivasan, Hansa and Selo, Vittorio and May, Rhys and Aisopos, Konstantinos and Hussenot, Léonard and Soares, Livio Baldini and Baumli, Kate and Chang, Michael B. and Recasens, Adrià and Caine, Ben and Pritzel, Alexander and Pavetic, Filip and Pardo, Fabio and Gergely, Anita and Frye, Justin and Ramasesh, Vinay and Horgan, Dan and Badola, Kartikeya and Kassner, Nora and Roy, Subhrajit and Dyer, Ethan and Campos, Víctor Campos and Tomala, Alex and Tang, Yunhao and Badawy, Dalia El and White, Elspeth and Mustafa, Basil and Lang, Oran and Jindal, Abhishek and Vikram, Sharad and Gong, Zhitao and Caelles, Sergi and Hemsley, Ross and Thornton, Gregory and Feng, Fangxiaoyu and Stokowiec, Wojciech and Zheng, Ce and Thacker, Phoebe and Ünlü, Çağlar and Zhang, Zhishuai and Saleh, Mohammad and Svensson, James and Bileschi, Max and Patil, Piyush and Anand, Ankesh and Ring, Roman and Tsihlas, Katerina and Vezer, Arpi and Selvi, Marco and Shevlane, Toby and Rodriguez, Mikel and Kwiatkowski, Tom and Daruki, Samira and Rong, Keran and Dafoe, Allan and FitzGerald, Nicholas and Gu-Lemberg, Keren and Khan, Mina and Hendricks, Lisa Anne and Pellat, Marie and Feinberg, Vladimir and Cobon-Kerr, James and Sainath, Tara and Rauh, Maribeth and Hashemi, Sayed Hadi and Ives, Richard and Hasson, Yana and Noland, Eric and Cao, Yuan and Byrd, Nathan and Hou, Le and Wang, Qingze and Sottiaux, Thibault and Paganini, Michela and Lespiau, Jean-Baptiste and Moufarek, Alexandre and Hassan, Samer and Shivakumar, Kaushik and Amersfoort, Joost van and Mandhane, Amol and Joshi, Pratik and Goyal, Anirudh and Tung, Matthew and Brock, Andrew and Sheahan, Hannah and Misra, Vedant and Li, Cheng and Rakićević, Nemanja and Dehghani, Mostafa and Liu, Fangyu and Mittal, Sid and Oh, Junhyuk and Noury, Seb and Sezener, Eren and Huot, Fantine and Lamm, Matthew and Cao, Nicola De and Chen, Charlie and Mudgal, Sidharth and Stella, Romina and Brooks, Kevin and Vasudevan, Gautam and Liu, Chenxi and Chain, Mainak and Melinkeri, Nivedita and Cohen, Aaron and Wang, Venus and Seymore, Kristie and Zubkov, Sergey and Goel, Rahul and Yue, Summer and Krishnakumaran, Sai and Albert, Brian and Hurley, Nate and Sano, Motoki and Mohananey, Anhad and Joughin, Jonah and Filonov, Egor and Kępa, Tomasz and Eldawy, Yomna and Lim, Jiawern and Rishi, Rahul and Badiezadegan, Shirin and Bos, Taylor and Chang, Jerry and Jain, Sanil and Padmanabhan, Sri Gayatri Sundara and Puttagunta, Subha and Krishna, Kalpesh and Baker, Leslie and Kalb, Norbert and Bedapudi, Vamsi and Kurzrok, Adam and Lei, Shuntong and Yu, Anthony and Litvin, Oren and Zhou, Xiang and Wu, Zhichun and Sobell, Sam and Siciliano, Andrea and Papir, Alan and Neale, Robby and Bragagnolo, Jonas and Toor, Tej and Chen, Tina and Anklin, Valentin and Wang, Feiran and Feng, Richie and Gholami, Milad and Ling, Kevin and Liu, Lijuan and Walter, Jules and Moghaddam, Hamid and Kishore, Arun and Adamek, Jakub and Mercado, Tyler and Mallinson, Jonathan and Wandekar, Siddhinita and Cagle, Stephen and Ofek, Eran and Garrido, Guillermo and Lombriser, Clemens and Mukha, Maksim and Sun, Botu and Mohammad, Hafeezul Rahman and Matak, Josip and Qian, Yadi and Peswani, Vikas and Janus, Pawel and Yuan, Quan and Schelin, Leif and David, Oana and Garg, Ankur and He, Yifan and Duzhyi, Oleksii and Älgmyr, Anton and Lottaz, Timothée and Li, Qi and Yadav, Vikas and Xu, Luyao and Chinien, Alex and Shivanna, Rakesh and Chuklin, Aleksandr and Li, Josie and Spadine, Carrie and Wolfe, Travis and Mohamed, Kareem and Das, Subhabrata and Dai, Zihang and He, Kyle and Dincklage, Daniel von and Upadhyay, Shyam and Maurya, Akanksha and Chi, Luyan and Krause, Sebastian and Salama, Khalid and Rabinovitch, Pam G. and M, Pavan Kumar Reddy and Selvan, Aarush and Dektiarev, Mikhail and Ghiasi, Golnaz and Guven, Erdem and Gupta, Himanshu and Liu, Boyi and Sharma, Deepak and Shtacher, Idan Heimlich and Paul, Shachi and Akerlund, Oscar and Aubet, François-Xavier and Huang, Terry and Zhu, Chen and Zhu, Eric and Teixeira, Elico and Fritze, Matthew and Bertolini, Francesco and Marinescu, Liana-Eleonora and Bölle, Martin and Paulus, Dominik and Gupta, Khyatti and Latkar, Tejasi and Chang, Max and Sanders, Jason and Wilson, Roopa and Wu, Xuewei and Tan, Yi-Xuan and Thiet, Lam Nguyen and Doshi, Tulsee and Lall, Sid and Mishra, Swaroop and Chen, Wanming and Luong, Thang and Benjamin, Seth and Lee, Jasmine and Andrejczuk, Ewa and Rabiej, Dominik and Ranjan, Vipul and Styrc, Krzysztof and Yin, Pengcheng and Simon, Jon and Harriott, Malcolm Rose and Bansal, Mudit and Robsky, Alexei and Bacon, Geoff and Greene, David and Mirylenka, Daniil and Zhou, Chen and Sarvana, Obaid and Goyal, Abhimanyu and Andermatt, Samuel and Siegler, Patrick and Horn, Ben and Israel, Assaf and Pongetti, Francesco and Chen, Chih-Wei "Louis" and Selvatici, Marco and Silva, Pedro and Wang, Kathie and Tolins, Jackson and Guu, Kelvin and Yogev, Roey and Cai, Xiaochen and Agostini, Alessandro and Shah, Maulik and Nguyen, Hung and Donnaile, Noah Ó and Pereira, Sébastien and Friso, Linda and Stambler, Adam and Kurzrok, Adam and Kuang, Chenkai and Romanikhin, Yan and Geller, Mark and Yan, Z. J. and Jang, Kane and Lee, Cheng-Chun and Fica, Wojciech and Malmi, Eric and Tan, Qijun and Banica, Dan and Balle, Daniel and Pham, Ryan and Huang, Yanping and Avram, Diana and Shi, Hongzhi and Singh, Jasjot and Hidey, Chris and Ahuja, Niharika and Saxena, Pranab and Dooley, Dan and Potharaju, Srividya Pranavi and O'Neill, Eileen and Gokulchandran, Anand and Foley, Ryan and Zhao, Kai and Dusenberry, Mike and Liu, Yuan and Mehta, Pulkit and Kotikalapudi, Ragha and Safranek-Shrader, Chalence and Goodman, Andrew and Kessinger, Joshua and Globen, Eran and Kolhar, Prateek and Gorgolewski, Chris and Ibrahim, Ali and Song, Yang and Eichenbaum, Ali and Brovelli, Thomas and Potluri, Sahitya and Lahoti, Preethi and Baetu, Cip and Ghorbani, Ali and Chen, Charles and Crawford, Andy and Pal, Shalini and Sridhar, Mukund and Gurita, Petru and Mujika, Asier and Petrovski, Igor and Cedoz, Pierre-Louis and Li, Chenmei and Chen, Shiyuan and Santo, Niccolò Dal and Goyal, Siddharth and Punjabi, Jitesh and Kappaganthu, Karthik and Kwak, Chester and LV, Pallavi and Velury, Sarmishta and Choudhury, Himadri and Hall, Jamie and Shah, Premal and Figueira, Ricardo and Thomas, Matt and Lu, Minjie and Zhou, Ting and Kumar, Chintu and Jurdi, Thomas and Chikkerur, Sharat and Ma, Yenai and Yu, Adams and Kwak, Soo and Ähdel, Victor and Rajayogam, Sujeevan and Choma, Travis and Liu, Fei and Barua, Aditya and Ji, Colin and Park, Ji Ho and Hellendoorn, Vincent and Bailey, Alex and Bilal, Taylan and Zhou, Huanjie and Khatir, Mehrdad and Sutton, Charles and Rzadkowski, Wojciech and Macintosh, Fiona and Vij, Roopali and Shagin, Konstantin and Medina, Paul and Liang, Chen and Zhou, Jinjing and Shah, Pararth and Bi, Yingying and Dankovics, Attila and Banga, Shipra and Lehmann, Sabine and Bredesen, Marissa and Lin, Zifan and Hoffmann, John Eric and Lai, Jonathan and Chung, Raynald and Yang, Kai and Balani, Nihal and Bražinskas, Arthur and Sozanschi, Andrei and Hayes, Matthew and Alcalde, Héctor Fernández and Makarov, Peter and Chen, Will and Stella, Antonio and Snijders, Liselotte and Mandl, Michael and Kärrman, Ante and Nowak, Paweł and Wu, Xinyi and Dyck, Alex and Vaidyanathan, Krishnan and R, Raghavender and Mallet, Jessica and Rudominer, Mitch and Johnston, Eric and Mittal, Sushil and Udathu, Akhil and Christensen, Janara and Verma, Vishal and Irving, Zach and Santucci, Andreas and Elsayed, Gamaleldin and Davoodi, Elnaz and Georgiev, Marin and Tenney, Ian and Hua, Nan and Cideron, Geoffrey and Leurent, Edouard and Alnahlawi, Mahmoud and Georgescu, Ionut and Wei, Nan and Zheng, Ivy and Scandinaro, Dylan and Jiang, Heinrich and Snoek, Jasper and Sundararajan, Mukund and Wang, Xuezhi and Ontiveros, Zack and Karo, Itay and Cole, Jeremy and Rajashekhar, Vinu and Tumeh, Lara and Ben-David, Eyal and Jain, Rishub and Uesato, Jonathan and Datta, Romina and Bunyan, Oskar and Wu, Shimu and Zhang, John and Stanczyk, Piotr and Zhang, Ye and Steiner, David and Naskar, Subhajit and Azzam, Michael and Johnson, Matthew and Paszke, Adam and Chiu, Chung-Cheng and Elias, Jaume Sanchez and Mohiuddin, Afroz and Muhammad, Faizan and Miao, Jin and Lee, Andrew and Vieillard, Nino and Park, Jane and Zhang, Jiageng and Stanway, Jeff and Garmon, Drew and Karmarkar, Abhijit and Dong, Zhe and Lee, Jong and Kumar, Aviral and Zhou, Luowei and Evens, Jonathan and Isaac, William and Irving, Geoffrey and Loper, Edward and Fink, Michael and Arkatkar, Isha and Chen, Nanxin and Shafran, Izhak and Petrychenko, Ivan and Chen, Zhe and Jia, Johnson and Levskaya, Anselm and Zhu, Zhenkai and Grabowski, Peter and Mao, Yu and Magni, Alberto and Yao, Kaisheng and Snaider, Javier and Casagrande, Norman and Palmer, Evan and Suganthan, Paul and Castaño, Alfonso and Giannoumis, Irene and Kim, Wooyeol and Rybiński, Mikołaj and Sreevatsa, Ashwin and Prendki, Jennifer and Soergel, David and Goedeckemeyer, Adrian and Gierke, Willi and Jafari, Mohsen and Gaba, Meenu and Wiesner, Jeremy and Wright, Diana Gage and Wei, Yawen and Vashisht, Harsha and Kulizhskaya, Yana and Hoover, Jay and Le, Maigo and Li, Lu and Iwuanyanwu, Chimezie and Liu, Lu and Ramirez, Kevin and Khorlin, Andrey and Cui, Albert and LIN, Tian and Wu, Marcus and Aguilar, Ricardo and Pallo, Keith and Chakladar, Abhishek and Perng, Ginger and Abellan, Elena Allica and Zhang, Mingyang and Dasgupta, Ishita and Kushman, Nate and Penchev, Ivo and Repina, Alena and Wu, Xihui and Weide, Tom van der and Ponnapalli, Priya and Kaplan, Caroline and Simsa, Jiri and Li, Shuangfeng and Dousse, Olivier and Yang, Fan and Piper, Jeff and Ie, Nathan and Pasumarthi, Rama and Lintz, Nathan and Vijayakumar, Anitha and Andor, Daniel and Valenzuela, Pedro and Lui, Minnie and Paduraru, Cosmin and Peng, Daiyi and Lee, Katherine and Zhang, Shuyuan and Greene, Somer and Nguyen, Duc Dung and Kurylowicz, Paula and Hardin, Cassidy and Dixon, Lucas and Janzer, Lili and Choo, Kiam and Feng, Ziqiang and Zhang, Biao and Singhal, Achintya and Du, Dayou and McKinnon, Dan and Antropova, Natasha and Bolukbasi, Tolga and Keller, Orgad and Reid, David and Finchelstein, Daniel and Raad, Maria Abi and Crocker, Remi and Hawkins, Peter and Dadashi, Robert and Gaffney, Colin and Franko, Ken and Bulanova, Anna and Leblond, Rémi and Chung, Shirley and Askham, Harry and Cobo, Luis C. and Xu, Kelvin and Fischer, Felix and Xu, Jun and Sorokin, Christina and Alberti, Chris and Lin, Chu-Cheng and Evans, Colin and Dimitriev, Alek and Forbes, Hannah and Banarse, Dylan and Tung, Zora and Omernick, Mark and Bishop, Colton and Sterneck, Rachel and Jain, Rohan and Xia, Jiawei and Amid, Ehsan and Piccinno, Francesco and Wang, Xingyu and Banzal, Praseem and Mankowitz, Daniel J. and Polozov, Alex and Krakovna, Victoria and Brown, Sasha and Bateni, MohammadHossein and Duan, Dennis and Firoiu, Vlad and Thotakuri, Meghana and Natan, Tom and Geist, Matthieu and Girgin, Ser tan and Li, Hui and Ye, Jiayu and Roval, Ofir and Tojo, Reiko and Kwong, Michael and Lee-Thorp, James and Yew, Christopher and Sinopalnikov, Danila and Ramos, Sabela and Mellor, John and Sharma, Abhishek and Wu, Kathy and Miller, David and Sonnerat, Nicolas and Vnukov, Denis and Greig, Rory and Beattie, Jennifer and Caveness, Emily and Bai, Libin and Eisenschlos, Julian and Korchemniy, Alex and Tsai, Tomy and Jasarevic, Mimi and Kong, Weize and Dao, Phuong and Zheng, Zeyu and Liu, Frederick and Yang, Fan and Zhu, Rui and Teh, Tian Huey and Sanmiya, Jason and Gladchenko, Evgeny and Trdin, Nejc and Toyama, Daniel and Rosen, Evan and Tavakkol, Sasan and Xue, Linting and Elkind, Chen and Woodman, Oliver and Carpenter, John and Papamakarios, George and Kemp, Rupert and Kafle, Sushant and Grunina, Tanya and Sinha, Rishika and Talbert, Alice and Wu, Diane and Owusu-Afriyie, Denese and Du, Cosmo and Thornton, Chloe and Pont-Tuset, Jordi and Narayana, Pradyumna and Li, Jing and Fatehi, Saaber and Wieting, John and Ajmeri, Omar and Uria, Benigno and Ko, Yeongil and Knight, Laura and Héliou, Amélie and Niu, Ning and Gu, Shane and Pang, Chenxi and Li, Yeqing and Levine, Nir and Stolovich, Ariel and Santamaria-Fernandez, Rebeca and Goenka, Sonam and Yustalim, Wenny and Strudel, Robin and Elqursh, Ali and Deck, Charlie and Lee, Hyo and Li, Zonglin and Levin, Kyle and Hoffmann, Raphael and Holtmann-Rice, Dan and Bachem, Olivier and Arora, Sho and Koh, Christy and Yeganeh, Soheil Hassas and Põder, Siim and Tariq, Mukarram and Sun, Yanhua and Ionita, Lucian and Seyedhosseini, Mojtaba and Tafti, Pouya and Liu, Zhiyu and Gulati, Anmol and Liu, Jasmine and Ye, Xinyu and Chrzaszcz, Bart and Wang, Lily and Sethi, Nikhil and Li, Tianrun and Brown, Ben and Singh, Shreya and Fan, Wei and Parisi, Aaron and Stanton, Joe and Koverkathu, Vinod and Choquette-Choo, Christopher A. and Li, Yunjie and Lu, T. J. and Ittycheriah, Abe and Shroff, Prakash and Varadarajan, Mani and Bahargam, Sanaz and Willoughby, Rob and Gaddy, David and Desjardins, Guillaume and Cornero, Marco and Robenek, Brona and Mittal, Bhavishya and Albrecht, Ben and Shenoy, Ashish and Moiseev, Fedor and Jacobsson, Henrik and Ghaffarkhah, Alireza and Rivière, Morgane and Walton, Alanna and Crepy, Clément and Parrish, Alicia and Zhou, Zongwei and Farabet, Clement and Radebaugh, Carey and Srinivasan, Praveen and Salm, Claudia van der and Fidjeland, Andreas and Scellato, Salvatore and Latorre-Chimoto, Eri and Klimczak-Plucińska, Hanna and Bridson, David and Cesare, Dario de and Hudson, Tom and Mendolicchio, Piermaria and Walker, Lexi and Morris, Alex and Mauger, Matthew and Guseynov, Alexey and Reid, Alison and Odoom, Seth and Loher, Lucia and Cotruta, Victor and Yenugula, Madhavi and Grewe, Dominik and Petrushkina, Anastasia and Duerig, Tom and Sanchez, Antonio and Yadlowsky, Steve and Shen, Amy and Globerson, Amir and Webb, Lynette and Dua, Sahil and Li, Dong and Bhupatiraju, Surya and Hurt, Dan and Qureshi, Haroon and Agarwal, Ananth and Shani, Tomer and Eyal, Matan and Khare, Anuj and Belle, Shreyas Rammohan and Wang, Lei and Tekur, Chetan and Kale, Mihir Sanjay and Wei, Jinliang and Sang, Ruoxin and Saeta, Brennan and Liechty, Tyler and Sun, Yi and Zhao, Yao and Lee, Stephan and Nayak, Pandu and Fritz, Doug and Vuyyuru, Manish Reddy and Aslanides, John and Vyas, Nidhi and Wicke, Martin and Ma, Xiao and Eltyshev, Evgenii and Martin, Nina and Cate, Hardie and Manyika, James and Amiri, Keyvan and Kim, Yelin and Xiong, Xi and Kang, Kai and Luisier, Florian and Tripuraneni, Nilesh and Madras, David and Guo, Mandy and Waters, Austin and Wang, Oliver and Ainslie, Joshua and Baldridge, Jason and Zhang, Han and Pruthi, Garima and Bauer, Jakob and Yang, Feng and Mansour, Riham and Gelman, Jason and Xu, Yang and Polovets, George and Liu, Ji and Cai, Honglong and Chen, Warren and Sheng, XiangHai and Xue, Emily and Ozair, Sherjil and Angermueller, Christof and Li, Xiaowei and Sinha, Anoop and Wang, Weiren and Wiesinger, Julia and Koukoumidis, Emmanouil and Tian, Yuan and Iyer, Anand and Gurumurthy, Madhu and Goldenson, Mark and Shah, Parashar and Blake, M. K. and Yu, Hongkun and Urbanowicz, Anthony and Palomaki, Jennimaria and Fernando, Chrisantha and Durden, Ken and Mehta, Harsh and Momchev, Nikola and Rahimtoroghi, Elahe and Georgaki, Maria and Raul, Amit and Ruder, Sebastian and Redshaw, Morgan and Lee, Jinhyuk and Zhou, Denny and Jalan, Komal and Li, Dinghua and Hechtman, Blake and Schuh, Parker and Nasr, Milad and Milan, Kieran and Mikulik, Vladimir and Franco, Juliana and Green, Tim and Nguyen, Nam and Kelley, Joe and Mahendru, Aroma and Hu, Andrea and Howland, Joshua and Vargas, Ben and Hui, Jeffrey and Bansal, Kshitij and Rao, Vikram and Ghiya, Rakesh and Wang, Emma and Ye, Ke and Sarr, Jean Michel and Preston, Melanie Moranski and Elish, Madeleine and Li, Steve and Kaku, Aakash and Gupta, Jigar and Pasupat, Ice and Juan, Da-Cheng and Someswar, Milan and M, Tejvi and Chen, Xinyun and Amini, Aida and Fabrikant, Alex and Chu, Eric and Dong, Xuanyi and Muthal, Amruta and Buthpitiya, Senaka and Jauhari, Sarthak and Hua, Nan and Khandelwal, Urvashi and Hitron, Ayal and Ren, Jie and Rinaldi, Larissa and Drath, Shahar and Dabush, Avigail and Jiang, Nan-Jiang and Godhia, Harshal and Sachs, Uli and Chen, Anthony and Fan, Yicheng and Taitelbaum, Hagai and Noga, Hila and Dai, Zhuyun and Wang, James and Liang, Chen and Hamer, Jenny and Ferng, Chun-Sung and Elkind, Chenel and Atias, Aviel and Lee, Paulina and Listík, Vít and Carlen, Mathias and Kerkhof, Jan van de and Pikus, Marcin and Zaher, Krunoslav and Müller, Paul and Zykova, Sasha and Stefanec, Richard and Gatsko, Vitaly and Hirnschall, Christoph and Sethi, Ashwin and Xu, Xingyu Federico and Ahuja, Chetan and Tsai, Beth and Stefanoiu, Anca and Feng, Bo and Dhandhania, Keshav and Katyal, Manish and Gupta, Akshay and Parulekar, Atharva and Pitta, Divya and Zhao, Jing and Bhatia, Vivaan and Bhavnani, Yashodha and Alhadlaq, Omar and Li, Xiaolin and Danenberg, Peter and Tu, Dennis and Pine, Alex and Filippova, Vera and Ghosh, Abhipso and Limonchik, Ben and Urala, Bhargava and Lanka, Chaitanya Krishna and Clive, Derik and Sun, Yi and Li, Edward and Wu, Hao and Hongtongsak, Kevin and Li, Ianna and Thakkar, Kalind and Omarov, Kuanysh and Majmundar, Kushal and Alverson, Michael and Kucharski, Michael and Patel, Mohak and Jain, Mudit and Zabelin, Maksim and Pelagatti, Paolo and Kohli, Rohan and Kumar, Saurabh and Kim, Joseph and Sankar, Swetha and Shah, Vineet and Ramachandruni, Lakshmi and Zeng, Xiangkai and Bariach, Ben and Weidinger, Laura and Vu, Tu and Andreev, Alek and He, Antoine and Hui, Kevin and Kashem, Sheleem and Subramanya, Amar and Hsiao, Sissie and Hassabis, Demis and Kavukcuoglu, Koray and Sadovsky, Adam and Le, Quoc and Strohman, Trevor and Wu, Yonghui and Petrov, Slav and Dean, Jeffrey and Vinyals, Oriol},
	month = may,
	year = {2025},
	note = {arXiv:2312.11805 [cs]},
}

@misc{zheng_judging_2023,
	title = {Judging {LLM}-as-a-{Judge} with {MT}-{Bench} and {Chatbot} {Arena}},
	url = {http://arxiv.org/abs/2306.05685},
	doi = {10.48550/arXiv.2306.05685},
	urldate = {2026-03-22},
	publisher = {arXiv},
	author = {Zheng, Lianmin and Chiang, Wei-Lin and Sheng, Ying and Zhuang, Siyuan and Wu, Zhanghao and Zhuang, Yonghao and Lin, Zi and Li, Zhuohan and Li, Dacheng and Xing, Eric P. and Zhang, Hao and Gonzalez, Joseph E. and Stoica, Ion},
	month = dec,
	year = {2023},
	note = {arXiv:2306.05685 [cs]},
}

@misc{joshi_triviaqa_2017,
	title = {{TriviaQA}: {A} {Large} {Scale} {Distantly} {Supervised} {Challenge} {Dataset} for {Reading} {Comprehension}},
	shorttitle = {{TriviaQA}},
	url = {http://arxiv.org/abs/1705.03551},
	doi = {10.48550/arXiv.1705.03551},
	urldate = {2026-03-17},
	publisher = {arXiv},
	author = {Joshi, Mandar and Choi, Eunsol and Weld, Daniel S. and Zettlemoyer, Luke},
	month = may,
	year = {2017},
	note = {arXiv:1705.03551 [cs]},
}

@article{hendrycks_measuring_2021,
	title = {Measuring {Massive} {Multitask} {Language} {Understanding}},
	journal = {Proceedings of the International Conference on Learning Representations (ICLR)},
	author = {Hendrycks, Dan and Burns, Collin and Basart, Steven and Zou, Andy and Mazeika, Mantas and Song, Dawn and Steinhardt, Jacob},
	year = {2021},
}

@misc{dua_drop_2019,
	title = {{DROP}: {A} {Reading} {Comprehension} {Benchmark} {Requiring} {Discrete} {Reasoning} {Over} {Paragraphs}},
	shorttitle = {{DROP}},
	url = {http://arxiv.org/abs/1903.00161},
	doi = {10.48550/arXiv.1903.00161},
	urldate = {2026-03-17},
	publisher = {arXiv},
	author = {Dua, Dheeru and Wang, Yizhong and Dasigi, Pradeep and Stanovsky, Gabriel and Singh, Sameer and Gardner, Matt},
	month = apr,
	year = {2019},
	note = {arXiv:1903.00161 [cs]
version: 2},
}

@article{lindsey_sparse_2024,
	title = {Sparse {Crosscoders} for {Cross}-{Layer} {Features} and {Model} {Diffing}},
	url = {https://transformer-circuits.pub/2024/crosscoders/index.html},
	journal = {Transformer Circuits Thread},
	author = {Lindsey, Jack and Templeton, Adly and Marcus, Josiah and Conerly, Tom and Batson, Joshua and Olah, Christopher},
	year = {2024},
}

@misc{wolf_huggingfaces_2019,
	title = {{HuggingFace}'s {Transformers}: {State}-of-the-art {Natural} {Language} {Processing}},
	shorttitle = {{HuggingFace}'s {Transformers}},
	url = {https://arxiv.org/abs/1910.03771v5},
	language = {en},
	urldate = {2026-03-12},
	journal = {arXiv.org},
	author = {Wolf, Thomas and Debut, Lysandre and Sanh, Victor and Chaumond, Julien and Delangue, Clement and Moi, Anthony and Cistac, Pierric and Rault, Tim and Louf, Rémi and Funtowicz, Morgan and Davison, Joe and Shleifer, Sam and von Platen, Patrick and Ma, Clara and Jernite, Yacine and Plu, Julien and Xu, Canwen and Scao, Teven Le and Gugger, Sylvain and Drame, Mariama and Lhoest, Quentin and Rush, Alexander M.},
	month = oct,
	year = {2019},
}

@misc{penedo_fineweb_2024,
	title = {The {FineWeb} {Datasets}: {Decanting} the {Web} for the {Finest} {Text} {Data} at {Scale}},
	shorttitle = {The {FineWeb} {Datasets}},
	url = {http://arxiv.org/abs/2406.17557},
	doi = {10.48550/arXiv.2406.17557},
	urldate = {2026-03-09},
	publisher = {arXiv},
	author = {Penedo, Guilherme and Kydlíček, Hynek and allal, Loubna Ben and Lozhkov, Anton and Mitchell, Margaret and Raffel, Colin and Werra, Leandro Von and Wolf, Thomas},
	month = oct,
	year = {2024},
	note = {arXiv:2406.17557 [cs]},
}

@misc{team_gemma_2025,
	title = {Gemma 3 {Technical} {Report}},
	url = {http://arxiv.org/abs/2503.19786},
	doi = {10.48550/arXiv.2503.19786},
	urldate = {2026-03-06},
	publisher = {arXiv},
	author = {Team, Gemma and Kamath, Aishwarya and Ferret, Johan and Pathak, Shreya and Vieillard, Nino and Merhej, Ramona and Perrin, Sarah and Matejovicova, Tatiana and Ramé, Alexandre and Rivière, Morgane and Rouillard, Louis and Mesnard, Thomas and Cideron, Geoffrey and Grill, Jean-bastien and Ramos, Sabela and Yvinec, Edouard and Casbon, Michelle and Pot, Etienne and Penchev, Ivo and Liu, Gaël and Visin, Francesco and Kenealy, Kathleen and Beyer, Lucas and Zhai, Xiaohai and Tsitsulin, Anton and Busa-Fekete, Robert and Feng, Alex and Sachdeva, Noveen and Coleman, Benjamin and Gao, Yi and Mustafa, Basil and Barr, Iain and Parisotto, Emilio and Tian, David and Eyal, Matan and Cherry, Colin and Peter, Jan-Thorsten and Sinopalnikov, Danila and Bhupatiraju, Surya and Agarwal, Rishabh and Kazemi, Mehran and Malkin, Dan and Kumar, Ravin and Vilar, David and Brusilovsky, Idan and Luo, Jiaming and Steiner, Andreas and Friesen, Abe and Sharma, Abhanshu and Sharma, Abheesht and Gilady, Adi Mayrav and Goedeckemeyer, Adrian and Saade, Alaa and Feng, Alex and Kolesnikov, Alexander and Bendebury, Alexei and Abdagic, Alvin and Vadi, Amit and György, András and Pinto, André Susano and Das, Anil and Bapna, Ankur and Miech, Antoine and Yang, Antoine and Paterson, Antonia and Shenoy, Ashish and Chakrabarti, Ayan and Piot, Bilal and Wu, Bo and Shahriari, Bobak and Petrini, Bryce and Chen, Charlie and Lan, Charline Le and Choquette-Choo, Christopher A. and Carey, C. J. and Brick, Cormac and Deutsch, Daniel and Eisenbud, Danielle and Cattle, Dee and Cheng, Derek and Paparas, Dimitris and Sreepathihalli, Divyashree Shivakumar and Reid, Doug and Tran, Dustin and Zelle, Dustin and Noland, Eric and Huizenga, Erwin and Kharitonov, Eugene and Liu, Frederick and Amirkhanyan, Gagik and Cameron, Glenn and Hashemi, Hadi and Klimczak-Plucińska, Hanna and Singh, Harman and Mehta, Harsh and Lehri, Harshal Tushar and Hazimeh, Hussein and Ballantyne, Ian and Szpektor, Idan and Nardini, Ivan and Pouget-Abadie, Jean and Chan, Jetha and Stanton, Joe and Wieting, John and Lai, Jonathan and Orbay, Jordi and Fernandez, Joseph and Newlan, Josh and Ji, Ju-yeong and Singh, Jyotinder and Black, Kat and Yu, Kathy and Hui, Kevin and Vodrahalli, Kiran and Greff, Klaus and Qiu, Linhai and Valentine, Marcella and Coelho, Marina and Ritter, Marvin and Hoffman, Matt and Watson, Matthew and Chaturvedi, Mayank and Moynihan, Michael and Ma, Min and Babar, Nabila and Noy, Natasha and Byrd, Nathan and Roy, Nick and Momchev, Nikola and Chauhan, Nilay and Sachdeva, Noveen and Bunyan, Oskar and Botarda, Pankil and Caron, Paul and Rubenstein, Paul Kishan and Culliton, Phil and Schmid, Philipp and Sessa, Pier Giuseppe and Xu, Pingmei and Stanczyk, Piotr and Tafti, Pouya and Shivanna, Rakesh and Wu, Renjie and Pan, Renke and Rokni, Reza and Willoughby, Rob and Vallu, Rohith and Mullins, Ryan and Jerome, Sammy and Smoot, Sara and Girgin, Sertan and Iqbal, Shariq and Reddy, Shashir and Sheth, Shruti and Põder, Siim and Bhatnagar, Sijal and Panyam, Sindhu Raghuram and Eiger, Sivan and Zhang, Susan and Liu, Tianqi and Yacovone, Trevor and Liechty, Tyler and Kalra, Uday and Evci, Utku and Misra, Vedant and Roseberry, Vincent and Feinberg, Vlad and Kolesnikov, Vlad and Han, Woohyun and Kwon, Woosuk and Chen, Xi and Chow, Yinlam and Zhu, Yuvein and Wei, Zichuan and Egyed, Zoltan and Cotruta, Victor and Giang, Minh and Kirk, Phoebe and Rao, Anand and Black, Kat and Babar, Nabila and Lo, Jessica and Moreira, Erica and Martins, Luiz Gustavo and Sanseviero, Omar and Gonzalez, Lucas and Gleicher, Zach and Warkentin, Tris and Mirrokni, Vahab and Senter, Evan and Collins, Eli and Barral, Joelle and Ghahramani, Zoubin and Hadsell, Raia and Matias, Yossi and Sculley, D. and Petrov, Slav and Fiedel, Noah and Shazeer, Noam and Vinyals, Oriol and Dean, Jeff and Hassabis, Demis and Kavukcuoglu, Koray and Farabet, Clement and Buchatskaya, Elena and Alayrac, Jean-Baptiste and Anil, Rohan and Dmitry and Lepikhin and Borgeaud, Sebastian and Bachem, Olivier and Joulin, Armand and Andreev, Alek and Hardin, Cassidy and Dadashi, Robert and Hussenot, Léonard},
	month = mar,
	year = {2025},
	note = {arXiv:2503.19786 [cs]},
}

@article{lindsey_biology_2025,
	title = {On the {Biology} of a {Large} {Language} {Model}},
	url = {https://transformer-circuits.pub/2025/attribution-graphs/biology.html},
	journal = {Transformer Circuits Thread},
	author = {Lindsey, Jack and Gurnee, Wes and Ameisen, Emmanuel and Chen, Brian and Pearce, Adam and Turner, Nicholas L. and Citro, Craig and Abrahams, David and Carter, Shan and Hosmer, Basil and Marcus, Jonathan and Sklar, Michael and Templeton, Adly and Bricken, Trenton and McDougall, Callum and Cunningham, Hoagy and Henighan, Thomas and Jermyn, Adam and Jones, Andy and Persic, Andrew and Qi, Zhenyi and Thompson, T. Ben and Zimmerman, Sam and Rivoire, Kelley and Conerly, Thomas and Olah, Chris and Batson, Joshua},
	year = {2025},
}

@misc{zellers_hellaswag_2019,
	title = {{HellaSwag}: {Can} a {Machine} {Really} {Finish} {Your} {Sentence}?},
	url = {https://arxiv.org/abs/1905.07830},
	author = {Zellers, Rowan and Holtzman, Ari and Bisk, Yonatan and Farhadi, Ali and Choi, Yejin},
	year = {2019},
	note = {\_eprint: 1905.07830},
}

@misc{zuo_moebert_2022,
	title = {{MoEBERT}: from {BERT} to {Mixture}-of-{Experts} via {Importance}-{Guided} {Adaptation}},
	url = {https://arxiv.org/abs/2204.07675},
	author = {Zuo, Simiao and Zhang, Qingru and Liang, Chen and He, Pengcheng and Zhao, Tuo and Chen, Weizhu},
	year = {2022},
	note = {\_eprint: 2204.07675},
}

@misc{romero_fitnets_2015,
	title = {{FitNets}: {Hints} for {Thin} {Deep} {Nets}},
	url = {https://arxiv.org/abs/1412.6550},
	author = {Romero, Adriana and Ballas, Nicolas and Kahou, Samira Ebrahimi and Chassang, Antoine and Gatta, Carlo and Bengio, Yoshua},
	year = {2015},
	note = {\_eprint: 1412.6550},
}

@inproceedings{durrani_analyzing_2020,
	address = {Online},
	title = {Analyzing {Individual} {Neurons} in {Pre}-trained {Language} {Models}},
	url = {https://aclanthology.org/2020.emnlp-main.395/},
	doi = {10.18653/v1/2020.emnlp-main.395},
	booktitle = {Proceedings of the 2020 {Conference} on {Empirical} {Methods} in {Natural} {Language} {Processing} ({EMNLP})},
	publisher = {Association for Computational Linguistics},
	author = {Durrani, Nadir and Sajjad, Hassan and Dalvi, Fahim and Belinkov, Yonatan},
	editor = {Webber, Bonnie and Cohn, Trevor and He, Yulan and Liu, Yang},
	month = nov,
	year = {2020},
	pages = {4865--4880},
}

@misc{hou_dynabert_2020,
	title = {{DynaBERT}: {Dynamic} {BERT} with {Adaptive} {Width} and {Depth}},
	url = {https://arxiv.org/abs/2004.04037},
	author = {Hou, Lu and Huang, Zhiqi and Shang, Lifeng and Jiang, Xin and Chen, Xiao and Liu, Qun},
	year = {2020},
	note = {\_eprint: 2004.04037},
}

@misc{sun_patient_2019,
	title = {Patient {Knowledge} {Distillation} for {BERT} {Model} {Compression}},
	url = {https://arxiv.org/abs/1908.09355},
	author = {Sun, Siqi and Cheng, Yu and Gan, Zhe and Liu, Jingjing},
	year = {2019},
	note = {\_eprint: 1908.09355},
}

@misc{dalvi_analyzing_2020,
	title = {Analyzing {Redundancy} in {Pretrained} {Transformer} {Models}},
	url = {https://arxiv.org/abs/2004.04010},
	author = {Dalvi, Fahim and Sajjad, Hassan and Durrani, Nadir and Belinkov, Yonatan},
	year = {2020},
	note = {\_eprint: 2004.04010},
}

@misc{antverg_pitfalls_2022,
	title = {On the {Pitfalls} of {Analyzing} {Individual} {Neurons} in {Language} {Models}},
	url = {https://arxiv.org/abs/2110.07483},
	author = {Antverg, Omer and Belinkov, Yonatan},
	year = {2022},
	note = {\_eprint: 2110.07483},
}

@misc{lee_self-knowledge_2022,
	title = {Self-{Knowledge} {Distillation} via {Dropout}},
	url = {https://arxiv.org/abs/2208.05642},
	author = {Lee, Hyoje and Park, Yeachan and Seo, Hyun and Kang, Myungjoo},
	year = {2022},
	note = {\_eprint: 2208.05642},
}

@misc{chen_symmetric_2017,
	title = {Symmetric {Variational} {Autoencoder} and {Connections} to {Adversarial} {Learning}},
	url = {https://arxiv.org/abs/1709.01846},
	author = {Chen, Liqun and Dai, Shuyang and Pu, Yunchen and Li, Chunyuan and Su, Qinliang and Carin, Lawrence},
	year = {2017},
	note = {\_eprint: 1709.01846},
}

@misc{nowozin_f-gan_2016,
	title = {f-{GAN}: {Training} {Generative} {Neural} {Samplers} using {Variational} {Divergence} {Minimization}},
	url = {https://arxiv.org/abs/1606.00709},
	author = {Nowozin, Sebastian and Cseke, Botond and Tomioka, Ryota},
	year = {2016},
	note = {\_eprint: 1606.00709},
}

@misc{huszar_how_2015,
	title = {How (not) to {Train} your {Generative} {Model}: {Scheduled} {Sampling}, {Likelihood}, {Adversary}?},
	shorttitle = {How (not) to {Train} your {Generative} {Model}},
	url = {https://arxiv.org/abs/1511.05101v1},
	language = {en},
	urldate = {2025-11-29},
	journal = {arXiv.org},
	author = {Huszár, Ferenc},
	month = nov,
	year = {2015},
}

@article{taori_alpaca_2023,
	title = {Alpaca: {A} strong, replicable instruction-following model},
	volume = {3},
	number = {6},
	journal = {Stanford Center for Research on Foundation Models. https://crfm. stanford. edu/2023/03/13/alpaca. html},
	author = {Taori, Rohan and Gulrajani, Ishaan and Zhang, Tianyi and Dubois, Yann and Li, Xuechen and Guestrin, Carlos and Liang, Percy and Hashimoto, Tatsunori B},
	year = {2023},
	pages = {7},
}

@misc{anthropic_get_2025,
	title = {Get started with {Claude}: {Quickstart}},
	url = {https://platform.claude.com/docs/en/get-started},
	author = {{Anthropic}},
	year = {2025},
	note = {Published: Claude Developer Platform Documentation},
}

@misc{google_gemini_2025,
	title = {Gemini {API} {Reference}},
	url = {https://ai.google.dev/api},
	author = {{Google}},
	year = {2025},
	note = {Published: Google AI for Developers},
}

@misc{openai_text_2025,
	title = {Text generation},
	url = {https://platform.openai.com/docs/guides/text},
	author = {{OpenAI}},
	year = {2025},
	note = {Published: OpenAI Platform Documentation},
}

@inproceedings{bucilua_model_2006,
	address = {New York, NY, USA},
	series = {{KDD} '06},
	title = {Model compression},
	isbn = {978-1-59593-339-3},
	url = {https://dl.acm.org/doi/10.1145/1150402.1150464},
	doi = {10.1145/1150402.1150464},
	urldate = {2025-11-28},
	booktitle = {Proceedings of the 12th {ACM} {SIGKDD} international conference on {Knowledge} discovery and data mining},
	publisher = {Association for Computing Machinery},
	author = {Buciluǎ, Cristian and Caruana, Rich and Niculescu-Mizil, Alexandru},
	month = aug,
	year = {2006},
	pages = {535--541},
}

@article{yang_survey_2024,
	title = {Survey on {Knowledge} {Distillation} for {Large} {Language} {Models}: {Methods}, {Evaluation}, and {Application}},
	issn = {2157-6904},
	shorttitle = {Survey on {Knowledge} {Distillation} for {Large} {Language} {Models}},
	url = {https://dl.acm.org/doi/10.1145/3699518},
	doi = {10.1145/3699518},
	urldate = {2025-10-29},
	journal = {ACM Trans. Intell. Syst. Technol.},
	author = {Yang, Chuanpeng and Zhu, Yao and Lu, Wang and Wang, Yidong and Chen, Qian and Gao, Chenlong and Yan, Bingjie and Chen, Yiqiang},
	month = oct,
	year = {2024},
	note = {Just Accepted},
}

@misc{jiao_tinybert_2020,
	title = {{TinyBERT}: {Distilling} {BERT} for {Natural} {Language} {Understanding}},
	url = {https://arxiv.org/abs/1909.10351},
	author = {Jiao, Xiaoqi and Yin, Yichun and Shang, Lifeng and Jiang, Xin and Chen, Xiao and Li, Linlin and Wang, Fang and Liu, Qun},
	year = {2020},
	note = {\_eprint: 1909.10351},
}

@misc{grattafiori_llama_2024,
	title = {The {Llama} 3 {Herd} of {Models}},
	url = {http://arxiv.org/abs/2407.21783},
	doi = {10.48550/arXiv.2407.21783},
	urldate = {2025-10-28},
	publisher = {arXiv},
	author = {Grattafiori, Aaron and Dubey, Abhimanyu and Jauhri, Abhinav and Pandey, Abhinav and Kadian, Abhishek and Al-Dahle, Ahmad and Letman, Aiesha and Mathur, Akhil and Schelten, Alan and Vaughan, Alex and Yang, Amy and Fan, Angela and Goyal, Anirudh and Hartshorn, Anthony and Yang, Aobo and Mitra, Archi and Sravankumar, Archie and Korenev, Artem and Hinsvark, Arthur and Rao, Arun and Zhang, Aston and Rodriguez, Aurelien and Gregerson, Austen and Spataru, Ava and Roziere, Baptiste and Biron, Bethany and Tang, Binh and Chern, Bobbie and Caucheteux, Charlotte and Nayak, Chaya and Bi, Chloe and Marra, Chris and McConnell, Chris and Keller, Christian and Touret, Christophe and Wu, Chunyang and Wong, Corinne and Ferrer, Cristian Canton and Nikolaidis, Cyrus and Allonsius, Damien and Song, Daniel and Pintz, Danielle and Livshits, Danny and Wyatt, Danny and Esiobu, David and Choudhary, Dhruv and Mahajan, Dhruv and Garcia-Olano, Diego and Perino, Diego and Hupkes, Dieuwke and Lakomkin, Egor and AlBadawy, Ehab and Lobanova, Elina and Dinan, Emily and Smith, Eric Michael and Radenovic, Filip and Guzmán, Francisco and Zhang, Frank and Synnaeve, Gabriel and Lee, Gabrielle and Anderson, Georgia Lewis and Thattai, Govind and Nail, Graeme and Mialon, Gregoire and Pang, Guan and Cucurell, Guillem and Nguyen, Hailey and Korevaar, Hannah and Xu, Hu and Touvron, Hugo and Zarov, Iliyan and Ibarra, Imanol Arrieta and Kloumann, Isabel and Misra, Ishan and Evtimov, Ivan and Zhang, Jack and Copet, Jade and Lee, Jaewon and Geffert, Jan and Vranes, Jana and Park, Jason and Mahadeokar, Jay and Shah, Jeet and Linde, Jelmer van der and Billock, Jennifer and Hong, Jenny and Lee, Jenya and Fu, Jeremy and Chi, Jianfeng and Huang, Jianyu and Liu, Jiawen and Wang, Jie and Yu, Jiecao and Bitton, Joanna and Spisak, Joe and Park, Jongsoo and Rocca, Joseph and Johnstun, Joshua and Saxe, Joshua and Jia, Junteng and Alwala, Kalyan Vasuden and Prasad, Karthik and Upasani, Kartikeya and Plawiak, Kate and Li, Ke and Heafield, Kenneth and Stone, Kevin and El-Arini, Khalid and Iyer, Krithika and Malik, Kshitiz and Chiu, Kuenley and Bhalla, Kunal and Lakhotia, Kushal and Rantala-Yeary, Lauren and Maaten, Laurens van der and Chen, Lawrence and Tan, Liang and Jenkins, Liz and Martin, Louis and Madaan, Lovish and Malo, Lubo and Blecher, Lukas and Landzaat, Lukas and Oliveira, Luke de and Muzzi, Madeline and Pasupuleti, Mahesh and Singh, Mannat and Paluri, Manohar and Kardas, Marcin and Tsimpoukelli, Maria and Oldham, Mathew and Rita, Mathieu and Pavlova, Maya and Kambadur, Melanie and Lewis, Mike and Si, Min and Singh, Mitesh Kumar and Hassan, Mona and Goyal, Naman and Torabi, Narjes and Bashlykov, Nikolay and Bogoychev, Nikolay and Chatterji, Niladri and Zhang, Ning and Duchenne, Olivier and Çelebi, Onur and Alrassy, Patrick and Zhang, Pengchuan and Li, Pengwei and Vasic, Petar and Weng, Peter and Bhargava, Prajjwal and Dubal, Pratik and Krishnan, Praveen and Koura, Punit Singh and Xu, Puxin and He, Qing and Dong, Qingxiao and Srinivasan, Ragavan and Ganapathy, Raj and Calderer, Ramon and Cabral, Ricardo Silveira and Stojnic, Robert and Raileanu, Roberta and Maheswari, Rohan and Girdhar, Rohit and Patel, Rohit and Sauvestre, Romain and Polidoro, Ronnie and Sumbaly, Roshan and Taylor, Ross and Silva, Ruan and Hou, Rui and Wang, Rui and Hosseini, Saghar and Chennabasappa, Sahana and Singh, Sanjay and Bell, Sean and Kim, Seohyun Sonia and Edunov, Sergey and Nie, Shaoliang and Narang, Sharan and Raparthy, Sharath and Shen, Sheng and Wan, Shengye and Bhosale, Shruti and Zhang, Shun and Vandenhende, Simon and Batra, Soumya and Whitman, Spencer and Sootla, Sten and Collot, Stephane and Gururangan, Suchin and Borodinsky, Sydney and Herman, Tamar and Fowler, Tara and Sheasha, Tarek and Georgiou, Thomas and Scialom, Thomas and Speckbacher, Tobias and Mihaylov, Todor and Xiao, Tong and Karn, Ujjwal and Goswami, Vedanuj and Gupta, Vibhor and Ramanathan, Vignesh and Kerkez, Viktor and Gonguet, Vincent and Do, Virginie and Vogeti, Vish and Albiero, Vítor and Petrovic, Vladan and Chu, Weiwei and Xiong, Wenhan and Fu, Wenyin and Meers, Whitney and Martinet, Xavier and Wang, Xiaodong and Wang, Xiaofang and Tan, Xiaoqing Ellen and Xia, Xide and Xie, Xinfeng and Jia, Xuchao and Wang, Xuewei and Goldschlag, Yaelle and Gaur, Yashesh and Babaei, Yasmine and Wen, Yi and Song, Yiwen and Zhang, Yuchen and Li, Yue and Mao, Yuning and Coudert, Zacharie Delpierre and Yan, Zheng and Chen, Zhengxing and Papakipos, Zoe and Singh, Aaditya and Srivastava, Aayushi and Jain, Abha and Kelsey, Adam and Shajnfeld, Adam and Gangidi, Adithya and Victoria, Adolfo and Goldstand, Ahuva and Menon, Ajay and Sharma, Ajay and Boesenberg, Alex and Baevski, Alexei and Feinstein, Allie and Kallet, Amanda and Sangani, Amit and Teo, Amos and Yunus, Anam and Lupu, Andrei and Alvarado, Andres and Caples, Andrew and Gu, Andrew and Ho, Andrew and Poulton, Andrew and Ryan, Andrew and Ramchandani, Ankit and Dong, Annie and Franco, Annie and Goyal, Anuj and Saraf, Aparajita and Chowdhury, Arkabandhu and Gabriel, Ashley and Bharambe, Ashwin and Eisenman, Assaf and Yazdan, Azadeh and James, Beau and Maurer, Ben and Leonhardi, Benjamin and Huang, Bernie and Loyd, Beth and Paola, Beto De and Paranjape, Bhargavi and Liu, Bing and Wu, Bo and Ni, Boyu and Hancock, Braden and Wasti, Bram and Spence, Brandon and Stojkovic, Brani and Gamido, Brian and Montalvo, Britt and Parker, Carl and Burton, Carly and Mejia, Catalina and Liu, Ce and Wang, Changhan and Kim, Changkyu and Zhou, Chao and Hu, Chester and Chu, Ching-Hsiang and Cai, Chris and Tindal, Chris and Feichtenhofer, Christoph and Gao, Cynthia and Civin, Damon and Beaty, Dana and Kreymer, Daniel and Li, Daniel and Adkins, David and Xu, David and Testuggine, Davide and David, Delia and Parikh, Devi and Liskovich, Diana and Foss, Didem and Wang, Dingkang and Le, Duc and Holland, Dustin and Dowling, Edward and Jamil, Eissa and Montgomery, Elaine and Presani, Eleonora and Hahn, Emily and Wood, Emily and Le, Eric-Tuan and Brinkman, Erik and Arcaute, Esteban and Dunbar, Evan and Smothers, Evan and Sun, Fei and Kreuk, Felix and Tian, Feng and Kokkinos, Filippos and Ozgenel, Firat and Caggioni, Francesco and Kanayet, Frank and Seide, Frank and Florez, Gabriela Medina and Schwarz, Gabriella and Badeer, Gada and Swee, Georgia and Halpern, Gil and Herman, Grant and Sizov, Grigory and Guangyi and Zhang and Lakshminarayanan, Guna and Inan, Hakan and Shojanazeri, Hamid and Zou, Han and Wang, Hannah and Zha, Hanwen and Habeeb, Haroun and Rudolph, Harrison and Suk, Helen and Aspegren, Henry and Goldman, Hunter and Zhan, Hongyuan and Damlaj, Ibrahim and Molybog, Igor and Tufanov, Igor and Leontiadis, Ilias and Veliche, Irina-Elena and Gat, Itai and Weissman, Jake and Geboski, James and Kohli, James and Lam, Janice and Asher, Japhet and Gaya, Jean-Baptiste and Marcus, Jeff and Tang, Jeff and Chan, Jennifer and Zhen, Jenny and Reizenstein, Jeremy and Teboul, Jeremy and Zhong, Jessica and Jin, Jian and Yang, Jingyi and Cummings, Joe and Carvill, Jon and Shepard, Jon and McPhie, Jonathan and Torres, Jonathan and Ginsburg, Josh and Wang, Junjie and Wu, Kai and U, Kam Hou and Saxena, Karan and Khandelwal, Kartikay and Zand, Katayoun and Matosich, Kathy and Veeraraghavan, Kaushik and Michelena, Kelly and Li, Keqian and Jagadeesh, Kiran and Huang, Kun and Chawla, Kunal and Huang, Kyle and Chen, Lailin and Garg, Lakshya and A, Lavender and Silva, Leandro and Bell, Lee and Zhang, Lei and Guo, Liangpeng and Yu, Licheng and Moshkovich, Liron and Wehrstedt, Luca and Khabsa, Madian and Avalani, Manav and Bhatt, Manish and Mankus, Martynas and Hasson, Matan and Lennie, Matthew and Reso, Matthias and Groshev, Maxim and Naumov, Maxim and Lathi, Maya and Keneally, Meghan and Liu, Miao and Seltzer, Michael L. and Valko, Michal and Restrepo, Michelle and Patel, Mihir and Vyatskov, Mik and Samvelyan, Mikayel and Clark, Mike and Macey, Mike and Wang, Mike and Hermoso, Miquel Jubert and Metanat, Mo and Rastegari, Mohammad and Bansal, Munish and Santhanam, Nandhini and Parks, Natascha and White, Natasha and Bawa, Navyata and Singhal, Nayan and Egebo, Nick and Usunier, Nicolas and Mehta, Nikhil and Laptev, Nikolay Pavlovich and Dong, Ning and Cheng, Norman and Chernoguz, Oleg and Hart, Olivia and Salpekar, Omkar and Kalinli, Ozlem and Kent, Parkin and Parekh, Parth and Saab, Paul and Balaji, Pavan and Rittner, Pedro and Bontrager, Philip and Roux, Pierre and Dollar, Piotr and Zvyagina, Polina and Ratanchandani, Prashant and Yuvraj, Pritish and Liang, Qian and Alao, Rachad and Rodriguez, Rachel and Ayub, Rafi and Murthy, Raghotham and Nayani, Raghu and Mitra, Rahul and Parthasarathy, Rangaprabhu and Li, Raymond and Hogan, Rebekkah and Battey, Robin and Wang, Rocky and Howes, Russ and Rinott, Ruty and Mehta, Sachin and Siby, Sachin and Bondu, Sai Jayesh and Datta, Samyak and Chugh, Sara and Hunt, Sara and Dhillon, Sargun and Sidorov, Sasha and Pan, Satadru and Mahajan, Saurabh and Verma, Saurabh and Yamamoto, Seiji and Ramaswamy, Sharadh and Lindsay, Shaun and Lindsay, Shaun and Feng, Sheng and Lin, Shenghao and Zha, Shengxin Cindy and Patil, Shishir and Shankar, Shiva and Zhang, Shuqiang and Zhang, Shuqiang and Wang, Sinong and Agarwal, Sneha and Sajuyigbe, Soji and Chintala, Soumith and Max, Stephanie and Chen, Stephen and Kehoe, Steve and Satterfield, Steve and Govindaprasad, Sudarshan and Gupta, Sumit and Deng, Summer and Cho, Sungmin and Virk, Sunny and Subramanian, Suraj and Choudhury, Sy and Goldman, Sydney and Remez, Tal and Glaser, Tamar and Best, Tamara and Koehler, Thilo and Robinson, Thomas and Li, Tianhe and Zhang, Tianjun and Matthews, Tim and Chou, Timothy and Shaked, Tzook and Vontimitta, Varun and Ajayi, Victoria and Montanez, Victoria and Mohan, Vijai and Kumar, Vinay Satish and Mangla, Vishal and Ionescu, Vlad and Poenaru, Vlad and Mihailescu, Vlad Tiberiu and Ivanov, Vladimir and Li, Wei and Wang, Wenchen and Jiang, Wenwen and Bouaziz, Wes and Constable, Will and Tang, Xiaocheng and Wu, Xiaojian and Wang, Xiaolan and Wu, Xilun and Gao, Xinbo and Kleinman, Yaniv and Chen, Yanjun and Hu, Ye and Jia, Ye and Qi, Ye and Li, Yenda and Zhang, Yilin and Zhang, Ying and Adi, Yossi and Nam, Youngjin and Yu and Wang and Zhao, Yu and Hao, Yuchen and Qian, Yundi and Li, Yunlu and He, Yuzi and Rait, Zach and DeVito, Zachary and Rosnbrick, Zef and Wen, Zhaoduo and Yang, Zhenyu and Zhao, Zhiwei and Ma, Zhiyu},
	month = nov,
	year = {2024},
	note = {arXiv:2407.21783 [cs]},
}

@misc{meta_ai_llama_2025,
	title = {The {Llama} 4 herd: {The} beginning of a new era of natively multimodal {AI} innovation},
	url = {https://ai.meta.com/blog/llama-4-multimodal-intelligence/},
	author = {{Meta AI}},
	month = apr,
	year = {2025},
}

@misc{team_kimi_2025,
	title = {Kimi {K2}: {Open} {Agentic} {Intelligence}},
	shorttitle = {Kimi {K2}},
	url = {http://arxiv.org/abs/2507.20534},
	doi = {10.48550/arXiv.2507.20534},
	urldate = {2025-10-28},
	publisher = {arXiv},
	author = {Team, Kimi and Bai, Yifan and Bao, Yiping and Chen, Guanduo and Chen, Jiahao and Chen, Ningxin and Chen, Ruijue and Chen, Yanru and Chen, Yuankun and Chen, Yutian and Chen, Zhuofu and Cui, Jialei and Ding, Hao and Dong, Mengnan and Du, Angang and Du, Chenzhuang and Du, Dikang and Du, Yulun and Fan, Yu and Feng, Yichen and Fu, Kelin and Gao, Bofei and Gao, Hongcheng and Gao, Peizhong and Gao, Tong and Gu, Xinran and Guan, Longyu and Guo, Haiqing and Guo, Jianhang and Hu, Hao and Hao, Xiaoru and He, Tianhong and He, Weiran and He, Wenyang and Hong, Chao and Hu, Yangyang and Hu, Zhenxing and Huang, Weixiao and Huang, Zhiqi and Huang, Zihao and Jiang, Tao and Jiang, Zhejun and Jin, Xinyi and Kang, Yongsheng and Lai, Guokun and Li, Cheng and Li, Fang and Li, Haoyang and Li, Ming and Li, Wentao and Li, Yanhao and Li, Yiwei and Li, Zhaowei and Li, Zheming and Lin, Hongzhan and Lin, Xiaohan and Lin, Zongyu and Liu, Chengyin and Liu, Chenyu and Liu, Hongzhang and Liu, Jingyuan and Liu, Junqi and Liu, Liang and Liu, Shaowei and Liu, T. Y. and Liu, Tianwei and Liu, Weizhou and Liu, Yangyang and Liu, Yibo and Liu, Yiping and Liu, Yue and Liu, Zhengying and Lu, Enzhe and Lu, Lijun and Ma, Shengling and Ma, Xinyu and Ma, Yingwei and Mao, Shaoguang and Mei, Jie and Men, Xin and Miao, Yibo and Pan, Siyuan and Peng, Yebo and Qin, Ruoyu and Qu, Bowen and Shang, Zeyu and Shi, Lidong and Shi, Shengyuan and Song, Feifan and Su, Jianlin and Su, Zhengyuan and Sun, Xinjie and Sung, Flood and Tang, Heyi and Tao, Jiawen and Teng, Qifeng and Wang, Chensi and Wang, Dinglu and Wang, Feng and Wang, Haiming and Wang, Jianzhou and Wang, Jiaxing and Wang, Jinhong and Wang, Shengjie and Wang, Shuyi and Wang, Yao and Wang, Yejie and Wang, Yiqin and Wang, Yuxin and Wang, Yuzhi and Wang, Zhaoji and Wang, Zhengtao and Wang, Zhexu and Wei, Chu and Wei, Qianqian and Wu, Wenhao and Wu, Xingzhe and Wu, Yuxin and Xiao, Chenjun and Xie, Xiaotong and Xiong, Weimin and Xu, Boyu and Xu, Jing and Xu, Jinjing and Xu, L. H. and Xu, Lin and Xu, Suting and Xu, Weixin and Xu, Xinran and Xu, Yangchuan and Xu, Ziyao and Yan, Junjie and Yan, Yuzi and Yang, Xiaofei and Yang, Ying and Yang, Zhen and Yang, Zhilin and Yang, Zonghan and Yao, Haotian and Yao, Xingcheng and Ye, Wenjie and Ye, Zhuorui and Yin, Bohong and Yu, Longhui and Yuan, Enming and Yuan, Hongbang and Yuan, Mengjie and Zhan, Haobing and Zhang, Dehao and Zhang, Hao and Zhang, Wanlu and Zhang, Xiaobin and Zhang, Yangkun and Zhang, Yizhi and Zhang, Yongting and Zhang, Yu and Zhang, Yutao and Zhang, Yutong and Zhang, Zheng and Zhao, Haotian and Zhao, Yikai and Zheng, Huabin and Zheng, Shaojie and Zhou, Jianren and Zhou, Xinyu and Zhou, Zaida and Zhu, Zhen and Zhuang, Weiyu and Zu, Xinxing},
	month = jul,
	year = {2025},
	note = {arXiv:2507.20534 [cs]},
}

@inproceedings{vaswani_attention_2017,
	title = {Attention is {All} you {Need}},
	volume = {30},
	url = {https://proceedings.neurips.cc/paper/2017/hash/3f5ee243547dee91fbd053c1c4a845aa-Abstract.html},
	urldate = {2025-10-28},
	booktitle = {Advances in {Neural} {Information} {Processing} {Systems}},
	publisher = {Curran Associates, Inc.},
	author = {Vaswani, Ashish and Shazeer, Noam and Parmar, Niki and Uszkoreit, Jakob and Jones, Llion and Gomez, Aidan N and Kaiser, Ł ukasz and Polosukhin, Illia},
	year = {2017},
}

@misc{hinton_distilling_2015,
	title = {Distilling the {Knowledge} in a {Neural} {Network}},
	url = {http://arxiv.org/abs/1503.02531},
	doi = {10.48550/arXiv.1503.02531},
	urldate = {2025-10-28},
	publisher = {arXiv},
	author = {Hinton, Geoffrey and Vinyals, Oriol and Dean, Jeff},
	month = mar,
	year = {2015},
	note = {arXiv:1503.02531 [stat]},
}

@article{achlioptas_database-friendly_2003,
	series = {Special {Issue} on {PODS} 2001},
	title = {Database-friendly random projections: {Johnson}-{Lindenstrauss} with binary coins},
	volume = {66},
	issn = {0022-0000},
	shorttitle = {Database-friendly random projections},
	url = {https://www.sciencedirect.com/science/article/pii/S0022000003000254},
	doi = {10.1016/S0022-0000(03)00025-4},
	number = {4},
	urldate = {2025-10-28},
	journal = {Journal of Computer and System Sciences},
	author = {Achlioptas, Dimitris},
	month = jun,
	year = {2003},
	pages = {671--687},
}

@misc{sanh_distilbert_2020,
	title = {{DistilBERT}, a distilled version of {BERT}: smaller, faster, cheaper and lighter},
	shorttitle = {{DistilBERT}, a distilled version of {BERT}},
	url = {http://arxiv.org/abs/1910.01108},
	doi = {10.48550/arXiv.1910.01108},
	urldate = {2025-10-13},
	publisher = {arXiv},
	author = {Sanh, Victor and Debut, Lysandre and Chaumond, Julien and Wolf, Thomas},
	month = mar,
	year = {2020},
	note = {arXiv:1910.01108 [cs]},
}

@inproceedings{wen_f-divergence_2023,
	address = {Toronto, Canada},
	title = {f-{Divergence} {Minimization} for {Sequence}-{Level} {Knowledge} {Distillation}},
	url = {https://aclanthology.org/2023.acl-long.605/},
	doi = {10.18653/v1/2023.acl-long.605},
	urldate = {2025-10-13},
	booktitle = {Proceedings of the 61st {Annual} {Meeting} of the {Association} for {Computational} {Linguistics} ({Volume} 1: {Long} {Papers})},
	publisher = {Association for Computational Linguistics},
	author = {Wen, Yuqiao and Li, Zichao and Du, Wenyu and Mou, Lili},
	editor = {Rogers, Anna and Boyd-Graber, Jordan and Okazaki, Naoaki},
	month = jul,
	year = {2023},
	pages = {10817--10834},
}

@misc{xu_survey_2024,
	title = {A {Survey} on {Knowledge} {Distillation} of {Large} {Language} {Models}},
	url = {http://arxiv.org/abs/2402.13116},
	doi = {10.48550/arXiv.2402.13116},
	urldate = {2025-10-13},
	publisher = {arXiv},
	author = {Xu, Xiaohan and Li, Ming and Tao, Chongyang and Shen, Tao and Cheng, Reynold and Li, Jinyang and Xu, Can and Tao, Dacheng and Zhou, Tianyi},
	month = oct,
	year = {2024},
	note = {arXiv:2402.13116 [cs]},
}

@article{gu_minillm_2024,
	title = {{MiniLLM}: {Knowledge} {Distillation} of {Large} {Language} {Models}},
	volume = {2024},
	shorttitle = {{MiniLLM}},
	url = {https://proceedings.iclr.cc/paper_files/paper/2024/hash/8ac015d409635f196f9e3e9dcfb9a94e-Abstract-Conference.html},
	language = {en},
	urldate = {2025-10-13},
	journal = {International Conference on Representation Learning},
	author = {Gu, Yuxian and Dong, Li and Wei, Furu and Huang, Minlie},
	month = may,
	year = {2024},
	pages = {32694--32717},
}

@article{clark_think_2018,
	title = {Think you have {Solved} {Question} {Answering}? {Try} {ARC}, the {AI2} {Reasoning} {Challenge}},
	journal = {arXiv:1803.05457v1},
	author = {Clark, Peter and Cowhey, Isaac and Etzioni, Oren and Khot, Tushar and Sabharwal, Ashish and Schoenick, Carissa and Tafjord, Oyvind},
	year = {2018},
}

@misc{sakaguchi_winogrande_2019,
	title = {{WinoGrande}: {An} {Adversarial} {Winograd} {Schema} {Challenge} at {Scale}},
	url = {https://arxiv.org/abs/1907.10641},
	author = {Sakaguchi, Keisuke and Bras, Ronan Le and Bhagavatula, Chandra and Choi, Yejin},
	year = {2019},
	note = {\_eprint: 1907.10641},
}

@misc{bisk_piqa_2019,
	title = {{PIQA}: {Reasoning} about {Physical} {Commonsense} in {Natural} {Language}},
	url = {https://arxiv.org/abs/1911.11641},
	author = {Bisk, Yonatan and Zellers, Rowan and Bras, Ronan Le and Gao, Jianfeng and Choi, Yejin},
	year = {2019},
	note = {\_eprint: 1911.11641},
}

@misc{nanda_transformerlens_2022,
	title = {{TransformerLens}},
	url = {https://github.com/TransformerLensOrg/TransformerLens},
	author = {Nanda, Neel and Bloom, Joseph},
	year = {2022},
}

@article{liang_homodistil_2023,
	title = {Homodistil: {Homotopic} task-agnostic distillation of pre-trained transformers},
	journal = {arXiv preprint arXiv:2302.09632},
	author = {Liang, Chen and Jiang, Haoming and Li, Zheng and Tang, Xianfeng and Yin, Bin and Zhao, Tuo},
	year = {2023},
}

@misc{wang_interpretability_2022,
	title = {Interpretability in the {Wild}: a {Circuit} for {Indirect} {Object} {Identification} in {GPT}-2 small},
	shorttitle = {Interpretability in the {Wild}},
	url = {http://arxiv.org/abs/2211.00593},
	doi = {10.48550/arXiv.2211.00593},
	urldate = {2025-01-07},
	publisher = {arXiv},
	author = {Wang, Kevin and Variengien, Alexandre and Conmy, Arthur and Shlegeris, Buck and Steinhardt, Jacob},
	month = nov,
	year = {2022},
	note = {arXiv:2211.00593 [cs]},
}

@misc{olsson_-context_2022,
	title = {In-context {Learning} and {Induction} {Heads}},
	url = {http://arxiv.org/abs/2209.11895},
	doi = {10.48550/arXiv.2209.11895},
	urldate = {2024-11-06},
	publisher = {arXiv},
	author = {Olsson, Catherine and Elhage, Nelson and Nanda, Neel and Joseph, Nicholas and DasSarma, Nova and Henighan, Tom and Mann, Ben and Askell, Amanda and Bai, Yuntao and Chen, Anna and Conerly, Tom and Drain, Dawn and Ganguli, Deep and Hatfield-Dodds, Zac and Hernandez, Danny and Johnston, Scott and Jones, Andy and Kernion, Jackson and Lovitt, Liane and Ndousse, Kamal and Amodei, Dario and Brown, Tom and Clark, Jack and Kaplan, Jared and McCandlish, Sam and Olah, Chris},
	month = sep,
	year = {2022},
	note = {arXiv:2209.11895},
}

@misc{elhage_toy_2022,
	title = {Toy {Models} of {Superposition}},
	url = {http://arxiv.org/abs/2209.10652},
	doi = {10.48550/arXiv.2209.10652},
	urldate = {2024-11-05},
	publisher = {arXiv},
	author = {Elhage, Nelson and Hume, Tristan and Olsson, Catherine and Schiefer, Nicholas and Henighan, Tom and Kravec, Shauna and Hatfield-Dodds, Zac and Lasenby, Robert and Drain, Dawn and Chen, Carol and Grosse, Roger and McCandlish, Sam and Kaplan, Jared and Amodei, Dario and Wattenberg, Martin and Olah, Christopher},
	month = sep,
	year = {2022},
	note = {arXiv:2209.10652},
}
\bibliographystyle{colm2026_conference}

\newpage

\appendix

\section{Disclosure of LLM Use}
We used LLMs to help generate the LaTeX code for tables and TikZ figures, citations for the developer APIs referenced in Section \ref{sss:whitevblack}, and Python code for the divergence comparison plot in Figure \ref{fig:jsd_kld_plot}.

\section{Extended Analysis Results}
\label{a:extended_analysis}

\begin{table}[htbp]
    \centering
    \small
    \begin{tabular}{lccccccccr}
        \textbf{Method} & \textbf{ARC} & \textbf{BoolQ} & \textbf{DROP} & \textbf{MMLU} & \textbf{PIQA} & \textbf{SciQ} & \textbf{TQA} & \textbf{WG} & \textbf{Avg.}\\
        \toprule
        KD   & \textit{0.8327} & 0.5936 & 0.0367 & 0.2319 & 0.7127 & 0.785 & 0.0388 & 0.5604 & 0.4232 \\
        LWD  & \textbf{\textit{0.8440}} & 0.6180 & 0.0442 & \textit{0.2322} & \textbf{\textit{0.7231}} & \textbf{\textit{0.799}} & 0.0622 & \textit{0.5675} & \textit{0.4303} \\
        TED  & 0.7818 & \textbf{\textit{0.6315}} & \textit{0.0320} & 0.2294 & \textit{0.7057} & 0.764 & 0.0154 & \textit{0.5667} & 0.4140 \\
        \midrule
        KDFP & 0.8407 & 0.6205 & \textbf{0.0447} & \textbf{0.2391} & 0.7165 & 0.786 & \textbf{0.0643} & \textbf{\textit{0.5722}} & \textbf{0.4316} \\
        \bottomrule
    \end{tabular}
    \caption{Results from finetuning teacher prior to distillation. Baseline and FT results are the same as in the Main Results in Table \ref{tab:main_results}, as there is no distillation. Italics indicate improvement over distillation without a finetuned teacher.}
    \label{tab:extended_ft_teacher}
\end{table}

\begin{table}[htbp]
    \centering
    \small
    \begin{tabular}{lccccccccr}
        \textbf{Method} & \textbf{ARC} & \textbf{BoolQ} & \textbf{DROP} & \textbf{MMLU} & \textbf{PIQA} & \textbf{SciQ} & \textbf{TQA} & \textbf{WG} & \textbf{Avg.}\\
        \toprule
        KD       & \textit{0.8327} & 0.5936 & 0.0367 & 0.2319 & \textbf{0.7231} & 0.785 & 0.0388 & \textbf{0.5675} & 0.4232 \\
        LWD      & \textit{0.8453} & 0.6080 & 0.0373 & \textbf{\textit{0.2352}} & 0.7165 & \textit{0.790} & 0.0595 & \textit{0.5635} & 0.4284 \\
        TED      & \textit{0.8119} & 0.6058 & 0.0237 & \textit{0.2313} & \textit{0.7046} & 0.765 & \textit{0.0485} & 0.5564 & \textit{0.4165} \\
        \midrule
        KDFP     & \textbf{0.8483} & \textbf{0.6144} & \textbf{0.0449} & 0.2339 & 0.7133 & \textbf{0.791} & \textbf{0.0600} & \textit{0.5667} & \textbf{0.4303} \\
        \bottomrule
    \end{tabular}
    \caption{Full results for the analysis of Pythia 6.9B as the teacher in the experiments in Section \ref{ss:big_teacher}. Italics indicate an improvement over the 2.8B parameter teacher in the Main Results in Table \ref{tab:main_results}.}
    \label{tab:extended_big_teacher}
\end{table}

\begin{table}[htbp]
    \centering
    \small
    \begin{tabular}{lccccccccr}
        \textbf{Method} & \textbf{ARC} & \textbf{BoolQ} & \textbf{DROP} & \textbf{MMLU} & \textbf{PIQA} & \textbf{SciQ} & \textbf{TQA} & \textbf{WG} & \textbf{Avg.}\\
        \toprule
        FT       & \textbf{\textit{0.8339}} & \textbf{\textit{0.6275}} & 0.0401 & \textbf{\textit{0.2407}} & \textit{0.7165} & 0.783 & \textit{0.0322} & 0.5627 & \textit{0.4263} \\
        KD       & 0.8136 & \textit{0.6177} & 0.0321 & \textit{0.2329} & 0.7155 & 0.774 & 0.0332 & \textbf{0.5675} & 0.4207 \\
        LWD      & 0.8313 & 0.6232 & 0.0391 & 0.2310 & 0.7155 & \textbf{\textit{0.790}} & \textbf{0.0517} & \textit{0.5572} & \textbf{0.4266} \\
        TED      & \textit{0.8080} & \textit{0.6168} & \textit{0.0257} & 0.2293 & 0.7013 & 0.762 & 0.0352 & \textit{0.5604} & 0.4154 \\
        \midrule
        KDFP     & 0.8326 & 0.6242 & \textbf{0.0416} & 0.2317 & \textbf{\textit{0.7209}} & 0.776 & 0.0499 & 0.5604 & 0.4264 \\
        \bottomrule
    \end{tabular}
    \caption{Full results for the data scaling analysis experiments in Section \ref{ss:5x_data}. Italics indicate an improvement over the 100M token trainig budget in the Main Results in Table \ref{tab:main_results}.}
    \label{tab:extended_5x_data}
\end{table}

\section{Incompatibility of Pretrained Student Attention Heads}
\label{a:incompat_attn_heads}
The specialization of attention heads presents an additional challenge for the distillation of pretrained students \citep{wang_interpretability_2022, olsson_-context_2022, elhage_mathematical_2021}. Since different attention heads autonomously specialize to perform different tasks \citep{wang_interpretability_2022, olsson_-context_2022, elhage_mathematical_2021}, it is very difficult to guarantee that attention heads are meaningfully distilled from the teacher to the student. While tools such as crosscoders \citep{lindsey_sparse_2024} may be capable of creating such a mapping, we leave it to future work and do not distill attention head activations.

\section{JS Divergence Implementation}
\label{a:js_details}
As is common in many KD methodologies, we employ a temperature-controlled softening of the output distributions with parameter $\tau$, where $\tau = 1$ returns the true predictions \citep{hinton_distilling_2015, jiao_tinybert_2020, liang_less_2023, xu_survey_2024}. We use the softmax function to compute the probability distribution from the raw logits.
\begin{align}
    p_i &= \frac{\exp(z_{T,i}/\tau)}{\sum_j \exp(z_{T,j}/\tau)}, \quad 
    q_i = \frac{\exp(z_{S,i}/\tau)}{\sum_j \exp(z_{S,j}/\tau)}
\end{align}

However, as shown in Equations \ref{eq:fwd_kld} and \ref{eq:beta-jsd}, we must compute the log probabilities, which can be numerically unstable if the softmax and log are applied sequentially. Instead, we use the log-sum-exp trick implemented in PyTorch's $\texttt{torch.nn.functional.log\_softmax}$ to directly compute the log probabilities, then exponentiate to recover the raw probabilities. 

Additionally, we exclude masked positions in the sequence from the loss calculation to ensure that loss is computed only over valid tokens. The final scalar loss is derived by averaging the element-wise JS divergence strictly over the count of unmasked tokens and scaled by $\tau^2$ to produce the final gradient, preventing masking and sequence padding from diluting the loss and accounting for the distribution softening.

\section{Projection Matrices}
\label{a:projection}
One of the critical challenges in distilling the internals of LLMs across models with different embedding dimensions is that a transformation is required to put the student and teacher activations into the same space for comparison. Traditionally, this has been done by adding a projection matrix\footnote{We use the terms \textit{projection matrix} and \textit{projector} interchangeably} $P$ to perform a linear transformation from the student space into the teacher space, which is learned during distillation \citep{liang_less_2023, jiao_tinybert_2020}.

Interestingly, this method of projection has not been tested against any alternatives. In fact, we did not find a single alternative to this projection formulation in our literature review. We hypothesize that learning a projection during training may \textit{dilute} the knowledge being transferred from teacher to student. The loss is computed in a shared space, but updates must propagate through the projector before it reaches the student in this formulation. It is possible that the projection matrix may \textit{absorb} some of the knowledge that is intended to be transferred to the student, only to be discarded after training. Additionally, this method does not consider projection in the opposite direction (i.e. projecting the teacher activations into the student space). We refer to the teacher-to-student direction as \textit{downward} and the reverse as \textit{upward}, to reflect the change in embedding dimension size.

We present a suite of projection formulations to compare the traditional approach alongside previously unexplored formulations, varying in their construction, training, and direction of projection.

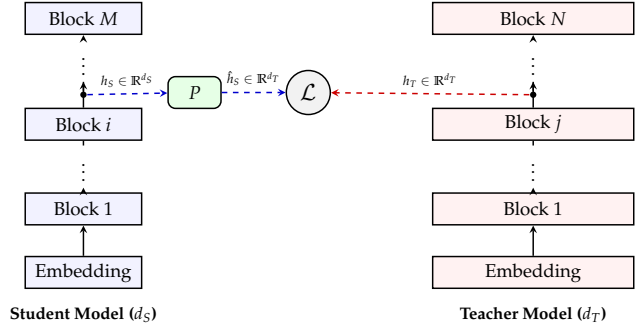
\begin{figure}[tbp]
    \centering
    \scalebox{0.7}{
    \begin{tikzpicture}[
        >=stealth,
        thick,
        box/.style={draw, rectangle, align=center, minimum height=0.6cm},
        student/.style={box, minimum width=2.2cm, fill=blue!5},
        teacher/.style={box, minimum width=3.8cm, fill=red!5},
        proj/.style={box, minimum width=1.0cm, fill=green!10, rounded corners},
        loss/.style={draw, circle, fill=gray!10, inner sep=4pt, font=\large},
        dot/.style={circle, fill, inner sep=1.2pt}
    ]
    
    \node[student] (s_emb) at (0, 0) {Embedding};
    \node[student] (s_b1) at (0, 1.2) {Block 1};
    \node (s_dots1) at (0, 2) {$\vdots$};
    \node[student] (s_bi) at (0, 2.8) {Block $i$};
    \node (s_dots2) at (0, 4.0) {$\vdots$};
    \node[student] (s_bM) at (0, 4.8) {Block $M$};
    
    \draw[->] (s_emb) -- (s_b1);
    \draw[->] (s_b1) -- (s_dots1);
    \draw[-]  (s_dots1) -- (s_bi);
    \draw[-]  (s_bi) -- (s_dots2) node[midway, dot] (s_tap) {};
    \draw[->] (s_tap) -- (s_dots2);
    \draw[->] (s_dots2) -- (s_bM);
    
    \node[teacher] (t_emb) at (8.5, 0) {Embedding};
    \node[teacher] (t_b1) at (8.5, 1.2) {Block 1};
    \node (t_dots1) at (8.5, 2) {$\vdots$};
    \node[teacher] (t_bj) at (8.5, 2.8) {Block $j$};
    \node (t_dots2) at (8.5, 4.0) {$\vdots$};
    \node[teacher] (t_bN) at (8.5, 4.8) {Block $N$};
    
    \draw[->] (t_emb) -- (t_b1);
    \draw[->] (t_b1) -- (t_dots1);
    \draw[-]  (t_dots1) -- (t_bj);
    \draw[-]  (t_bj) -- (t_dots2) node[midway, dot] (t_tap) {};
    \draw[->] (t_tap) -- (t_dots2);
    \draw[->] (t_dots2) -- (t_bN);
    
    \node[loss] (l) at (4.25, 3.4) {$\mathcal{L}$};
    \node[proj] (p) at (2.1, 3.4) {$P$};
    
    \draw[->, dashed, blue!80!black] (s_tap) -- (p) 
        node[midway, above, font=\scriptsize, text=black] {$h_S \in \mathbb{R}^{d_S}$};
    
    \draw[->, dashed, blue!80!black] (p) -- (l) 
        node[midway, above, font=\scriptsize, text=black] {$\hat{h}_S \in \mathbb{R}^{d_T}$};
    
    \draw[->, dashed, red!80!black] (t_tap) -- (l) 
        node[midway, above, font=\scriptsize, text=black] {$h_T \in \mathbb{R}^{d_T}$};
    
    \node[below=0.2cm of s_emb, font=\small\bfseries] {Student Model ($d_S$)};
    \node[below=0.2cm of t_emb, font=\small\bfseries] {Teacher Model ($d_T$)};
    
    \end{tikzpicture}
    }
    \caption{Distillation of internal activations between a student and teacher transformer with an upward projection. The output of the residual stream after student Block $i$ ($h_S$) is passed through a projection matrix $P$. This maps the student's smaller embedding dimension ($d_S$) into the teacher's latent space ($d_T$) so that it can be directly compared against the target teacher activation ($h_T$) from Block $j$.}
    \label{fig:transformer_kd}
\end{figure}

\subsection{Online}
The \textit{online} projection matrices formed using the existing method of learning the weights during distillation. Although simple to implement, it does not necessarily train students to learn representations which align with those of the teacher. Since the alignment inside of the loss computation is a product of the projection matrix, it is possible that online projector captures the necessary alignment knowledge instead of propagating that knowledge into the student model.

\subsection{Offline}
\label{as:off_proj}

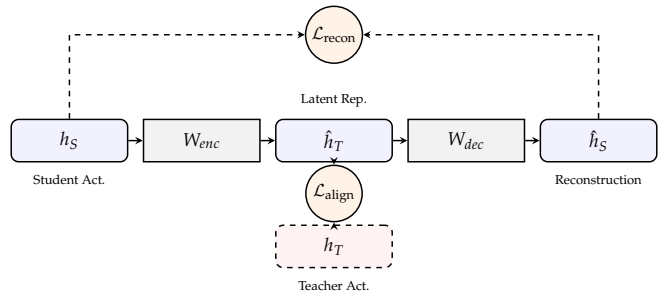
\begin{figure}[htbp]
\centering
\scalebox{0.7}{
\begin{tikzpicture}[
    >=stealth, thick,
    box/.style={draw, rectangle, minimum height=0.8cm, align=center},
    data/.style={box, rounded corners, fill=blue!5, minimum width=2.2cm},
    target/.style={box, rounded corners, fill=red!5, dashed, minimum width=2.2cm},
    layer/.style={box, fill=gray!10, minimum width=2.2cm},
    loss/.style={draw, circle, fill=orange!10, inner sep=2pt, font=\small}
]

\node[data] (input) at (0,0) {$h_S$};
\node[layer] (enc) at (2.5,0) {$W_{enc}$};
\node[data] (latent) at (5,0) {$\hat{h}_T$};
\node[layer] (dec) at (7.5,0) {$W_{dec}$};
\node[data] (output) at (10,0) {$\hat{h}_S$};

\node[target] (teacher) at (5,-2) {$h_T$};

\node[loss] (l_align) at (5,-1) {$\mathcal{L}_{\text{align}}$};
\node[loss] (l_recon) at (5, 2) {$\mathcal{L}_{\text{recon}}$};

\draw[->] (input) -- (enc);
\draw[->] (enc) -- (latent);
\draw[->] (latent) -- (dec);
\draw[->] (dec) -- (output);

\draw[->, dashed] (latent) -- (l_align);
\draw[->, dashed] (teacher) -- (l_align);

\draw[->, dashed] (input) |- (l_recon);
\draw[->, dashed] (output) |- (l_recon);

\node[below=0.1cm of input, font=\scriptsize] {Student Act.};
\node[below=0.1cm of output, font=\scriptsize] {Reconstruction};
\node[above=0.1cm of latent, font=\scriptsize] {Latent Rep.};
\node[below=0.1cm of teacher, font=\scriptsize] {Teacher Act.};

\end{tikzpicture}
}
\caption{The fully linear autoencoder architecture for an upward offline projector. The student activations ($x = h_S$) serve as the input and are reconstructed as $\hat{h}_S$ via linear weights $W_{enc}$ and $W_{dec}$. The intermediate latent representation ($\hat{z} = \hat{h}_T$) is explicitly trained to match the teacher activations ($z = h_T$). The total objective is a convex combination of the reconstruction loss ($\mathcal{L}_{\text{recon}}$) and the latent alignment loss ($\mathcal{L}_{\text{align}}$).}
\label{fig:off_proj_diagram}
\end{figure}

We use the term \textit{offline} projector to refer to projection matrices trained separately prior to distillation, then inserted and frozen as the student is trained. Offline projectors are trained as fully linear autoencoders on activations obtained from the student and teacher models. 
\begin{align}
    \hat{z} &= W_{enc}(x) \\
    \hat{x} &= W_{dec}(\hat{z})
\end{align}

We train the offline projectors using a combination of a reconstruction loss on the output, $\hat{x} - x$ and an alignment loss on the hidden representations, $\hat{z} - z$. The parameter $\beta$ is used to tune the weighting of the alignment loss.
\begin{align}
    \mathcal{L}_{rec}(x) &= MSE(\hat{x} - x) \\ 
    \mathcal{L}_{align}(x) &= MSE(\hat{z} - z) \\
    \mathcal{L}_{total}(x) &= \beta \cdot \mathcal{L}_{align}(x) \\ \nonumber
    &+ (1-\beta) \cdot \mathcal{L}_{rec}(x)
\end{align}

Our motivation for the offline approach is this: if the model activations fall into a particular subspace of the full embedding space, then an autoencoder trained on simple reconstruction would not be able to account for this. Thus, the additional loss on the latent representations of the autoencoder serve to align the generated latent representations with \textit{actual} collected representations. We collected activations from 10M tokens of the FineWeb Edu dataset \citep{penedo_fineweb_2024} for Pythia 2.8b deduped (teacher) and Pythia 1.4b deduped (student). We train each indivdual offline projectors using activations at \textit{resid\_post}, \textit{attn\_out}, and \textit{mlp\_out} hook points at every layer of the student model, and at linearly spaced layers of the teacher model such that a one-to-one mapping exists between selected student and teacher layers. We experimented with latent alignment parameters of 0 (no alignment loss), 0.25, 0.5, 0.75, and 1 (no reconstruction loss).

\subsection{Randomized}
The \textit{randomized} projection matrix is created using the binary coin formulation from \citet{achlioptas_database-friendly_2003} and frozen during training. The theoretical guarantees regarding preservation of distances between points only hold in the downward projection (i.e. projecting teacher activations into the student space). In support of our effort to conduct a thorough empirical evaluation, we run experiments on upward randomized projections, despite the lack of theoretical motivation.

\section{Exploration Process}
\label{a:exploration}
The KDFP methodology was finalized as the culmination of a multi-stage exploration. In all stages, we use Pythia 2.8b deduped as the teacher, Pythia 1.4b as the pretrained student base model, 10m tokens of FineWeb Edu to train the offline projectors, 100m tokens of FineWeb Edu to train the students, and enforce reproducibility with a seed of 3. An input sequence length of 2048 was used for the entirety of the exploration, as this is the maximum context size for Pythia models \citep{biderman_pythia_2023}. We evaluate each configuration on HellaSwag accuracy, using 0-shot evaluations unless otherwise specified below \citep{zellers_hellaswag_2019}. Since our exploration is essentially a parametrized system, the results in the main body of the paper do not include performance on HellaSwag so as not to introduce bias into the results.

We conducted 5 stages of exploration:
\begin{enumerate}
    \item Basic Exploration
    \item Search for Offline Projector Alignment
    \item Search for Divergence Parameter
    \item Search for Objective Function
    \item Final Validation
\end{enumerate}

\begin{figure}[tbp]
    \centering
    \label{fig:student-architectures}
    \scalebox{0.7}{
    \begin{tikzpicture}[
        node distance=1.2cm,
        layer/.style={rectangle, draw=black!80, line width=0.7pt, font=\small, minimum height=0.5cm},
        teacher/.style={fill=blue!10, minimum width=3.2cm},
        shallow/.style={fill=orange!10, minimum width=3.2cm},
        generic/.style={fill=green!10, minimum width=1.6cm},
        arrow/.style={-stealth, thick, black!60},
        skip_arrow/.style={-stealth, thick, black!60, densely dotted},
        width_brace/.style={decorate, decoration={brace, amplitude=4pt}, thick, gray!80}
    ]
    
        \node[layer, teacher] (t_in) {Input};
        \node[layer, teacher, above=0.3cm of t_in] (t_1) {Layer $1$};
        \node[layer, teacher, above=1.2cm of t_1] (t_N) {Layer $N$};
        \node[layer, teacher, above=0.3cm of t_N] (t_out) {Output};
        
        \draw[arrow] (t_in) -- (t_1);
        \draw[skip_arrow] (t_1) -- (t_N);
        \draw[arrow] (t_N) -- (t_out);
        
        \node[below=0.2cm of t_in, font=\bfseries] {Teacher};
        \node[below=0.6cm of t_in, font=\footnotesize] {($N$ Layers)};
    
        \draw[width_brace] (t_out.north west) -- (t_out.north east) node[midway, above=0.15cm, font=\scriptsize, text=black!80] {Width $W$};
    
        \node[layer, shallow, right=1.5cm of t_in] (s_in) {Input};
        \node[layer, shallow, above=0.3cm of s_in] (s_1) {Layer $1$};
        \node[layer, shallow, above=0.4cm of s_1] (s_Nk) {Layer $N/k$};
        \node[layer, shallow, above=0.3cm of s_Nk] (s_out) {Output};
        
        \draw[arrow] (s_in) -- (s_1);
        \draw[skip_arrow] (s_1) -- (s_Nk);
        \draw[arrow] (s_Nk) -- (s_out);
        
        \node[below=0.2cm of s_in, font=\bfseries] {Shallow Student};
        \node[below=0.6cm of s_in, font=\footnotesize] {($N/k$ Layers)};
    
        \draw[width_brace] (s_out.north west) -- (s_out.north east) node[midway, above=0.15cm, font=\scriptsize, text=black!80] {Width $W$};
    
        \node[layer, generic, right=1.5cm of s_in] (g_in) {Input};
        \node[layer, generic, above=0.3cm of g_in] (g_1) {Layer $1$};
        \node[layer, generic, above=0.8cm of g_1] (g_P) {Layer $P$};
        \node[layer, generic, above=0.3cm of g_P] (g_out) {Output};
        
        \draw[arrow] (g_in) -- (g_1);
        \draw[skip_arrow] (g_1) -- (g_P);
        \draw[arrow] (g_P) -- (g_out);
        
        \node[below=0.2cm of g_in, font=\bfseries] {Pretrained Student};
        \node[below=0.6cm of g_in, font=\footnotesize] {($P$ Layers)};
    
        \draw[width_brace] (g_out.north west) -- (g_out.north east) node[midway, above=0.15cm, font=\scriptsize, text=black!80] {Width $w < W$};
    \end{tikzpicture}
    }
    \caption{Shallow student architectures used to be a common choice in KD, as it is maximally compatible with the teacher. However, shallow students require full pretraining, making them infeasible in most modern settings. Selecting a pretrained model for student initialization is the current best practice for continual pretraining.}
\end{figure}
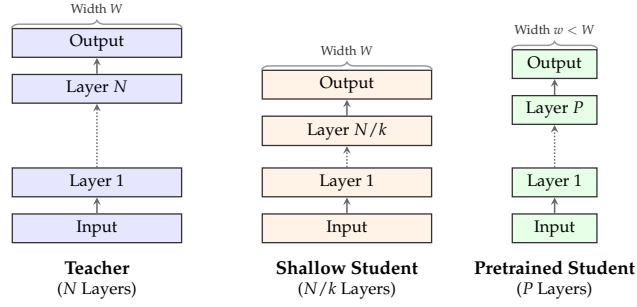

\subsection{Stage 1: Basic Exploration}
\label{ss:stage1}
We ran both pretrained and shallow student configurations with all projector variants in each direction. All objectives utilized a balanced objective (i.e. equal weighting across all available components) with true JS divergence ($\beta = 0.5$). Offline projectors were trained with an alignment parameter of 0.5. The goal of stage one was primarily to get a feel for how each of the different KD configurations performed with sensible defaults. The two notable findings were that shallow students were far inferior to pretrained, and that projection direction made no difference for offline and randomized projections, and only a very small difference for online. With such poor initial performance from shallow students, we chose not to include them in any further stages of exploration.

\begin{table}[htbp]
\centering
\begin{tabular}{lc}
\hline
\textbf{Experiment} & \textbf{Accuracy} \\ \hline
pretrained-ft                                            & 0.4146 \\
pretrained-no\_train                                      & 0.4164 \\
pretrained-kd-down\_offline & 0.4167 \\
pretrained-kd-up\_offline   & 0.4167 \\
pretrained-kd-down\_randomized       & 0.4167 \\
pretrained-kd-up\_randomized           & 0.4167 \\
pretrained-kd-down\_online             & 0.4131 \\
pretrained-kd-up\_online               & 0.4126 \\ \hline
shallow-ft                                               & 0.3016 \\
shallow-no\_train                                        & 0.2617 \\
shallow-kd                            & 0.2590 \\ \hline
\end{tabular}
\caption{Stage 1: Results of the basic exploration}
\label{tab:stage1}
\end{table}

\subsection{Stage 2: Search for Offline Projector Alignment}
\label{ss:stage2}
In our experiments exploring the effects of different offline projector alignment parameters, we found that every variant achieved an identical performance of 0.4151 on 0-shot HellaSwag evaluations. This included a full sweep of all alignment parameter values, in both upward and downward projection directions. These experiments were run with a balanced objective and regular JS Divergence ($\beta$ = 0.5). 

Given this result, we fixed the alignment parameter to 0, as this simplifies the training of the offline projector by only requiring activations from one of the models. We also chose to fix the offline projector as an upward projection. The upward projection preserves the full fidelity of the teacher activations- which serve as the ground truth in KD- and excluding latent alignment simplifies the training of the offline projector by removing the need for an additional set of model activations. To reduce the search space, we proceeded with upward projections as the selected direction for all projections, and left further evaluation of downward projections to the final stage.

\subsection{Stage 3: Search for Divergence Parameter}
\label{a:div_search}
We experimented with the parametrized JS divergence using $\beta$ = 0, 0.25, 0.5, 0.75, and 1. $\beta$ = 0 corresponds to forward KL divergence, $\beta$ = 1 to reverse KL divergence, and $\beta$ = 0.5 is the traditional JS divergence. Since the divergence measure is applied only to the output distributions of the teacher and student, we chose to isolate the effects of the divergence parameter by only including the output distributions and the true labels in objective function. Note that the selected projection is irrelevant, as none of the projected activations are included in the objective. We started with 0-shot evaluations but did not find as large of a gap as we hypothesized, and thus also ran 5-shot. We found inconsistency in the top and bottom performing parameters across the two sets of evaluations. We selected $\beta$ = 0.25 as our divergence parameter, as it was the second highest performer in both evaluations.

\begin{table}[htbp]
\centering
\begin{tabular}{ccc}
\toprule
& \multicolumn{2}{c}{\textbf{Accuracy}} \\
\cmidrule(lr){2-3}
\textbf{$\beta$} & \textbf{0-shot} & \textbf{5-shot} \\
\midrule
0.0  & 0.4150 & 0.4132 \\
0.25 & 0.4151 & 0.4133 \\
0.5  & 0.4148 & \underline{0.4128} \\
0.75 & \textbf{0.4156} & \underline{0.4128} \\
1.0  & \underline{0.4147} & \textbf{0.4134} \\
\bottomrule
\end{tabular}
\caption{Accuracy on HellaSwag across different divergence parameter values for 0-shot and 5-shot evaluations. Best results are bolded, and the lowest are underlined.}
\label{tab:stage3}
\end{table}

\subsection{Stage 4: Search for Objective Function}
\label{a:objective}
Given the large search space for possible objective functions which could be formed from the logits and activations we targeted, we had to take measures to make the search feasible. We set the constraint that every objective function is a convex combination of all of the included components, with weights evenly distributed. We also used a two-part approach to reduce the number of combinations of objective components to explore.

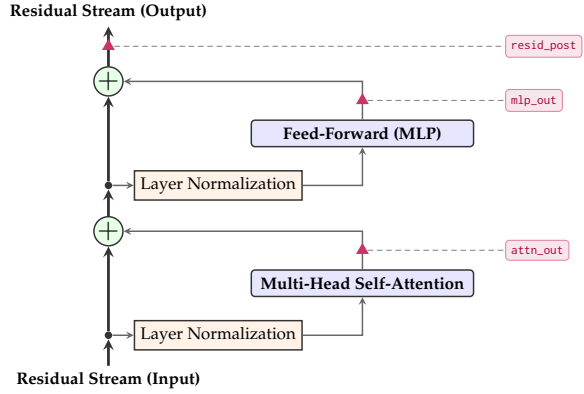
\begin{figure}[tbp]
    \centering
    \scalebox{0.7}{
        \begin{tikzpicture}[
        node distance=1.2cm,
        block/.style={rectangle, draw=black!80, fill=blue!10, line width=0.8pt, font=\small\bfseries, minimum height=0.5cm, minimum width=4.2cm, rounded corners=2pt, align=center},
        norm/.style={rectangle, draw=black!80, fill=orange!10, line width=0.8pt, font=\small, minimum height=0.5cm, minimum width=2.8cm, align=center},
        add/.style={circle, draw=black!80, fill=green!10, line width=0.8pt, font=\Large\bfseries, minimum size=0.25cm, inner sep=0pt},
        %
        main_stream/.style={-stealth, line width=1.5pt, black!80}, 
        branch_stream/.style={-stealth, thick, black!60},
        split_dot/.style={circle, fill=black!80, inner sep=0pt, minimum size=4pt},
        %
        hook/.style={rectangle, draw=purple!50, fill=purple!10, text=purple!90!black, font=\ttfamily\scriptsize, rounded corners=2pt, inner sep=3pt},
        hook_tri/.style={regular polygon, regular polygon sides=3, fill=purple!80, inner sep=0pt, minimum size=8pt}
    ]
    
        \coordinate (label_col) at (7.5, 0);
    
        \coordinate (block_center) at (4.8, 0);
    
        \node[font=\small\bfseries] (input) {Residual Stream (Input)};
        
        \node[split_dot, above=0.5cm of input] (split1) {};
        \node[add, above=1.6cm of split1] (add1) {$+$};
        
        \node[split_dot, above=0.5cm of add1] (split2) {};
        \node[add, above=1.6cm of split2] (add2) {$+$};
        
        \node[font=\small\bfseries, above=0.75cm of add2] (output) {Residual Stream (Output)};
    
        
        \node[norm, right=0.4cm of split1] (norm1) {Layer Normalization};
        \node[block] (mha) at (block_center |- {$(split1)!0.5!(add1)$}) {Multi-Head Self-Attention};
    
        \node[norm, right=0.4cm of split2] (norm2) {Layer Normalization};
        \node[block] (mlp) at (block_center |- {$(split2)!0.5!(add2)$}) {Feed-Forward (MLP)};
    
        
        \draw[main_stream] (input) -- (split1);
        \draw[main_stream] (split1) -- (add1);
        \draw[main_stream] (add1) -- (split2);
        \draw[main_stream] (split2) -- (add2);
        
        \draw[main_stream] (add2) -- node[hook_tri] (resid_tri) {} (output);
    
        \draw[branch_stream] (split1) -- (norm1.west);
        \draw[branch_stream] (norm1.east) -| (mha.south);
        \draw[branch_stream] (mha.north) -- node[hook_tri] (attn_tri) {} (mha.north |- add1.east) -- (add1.east); 
    
        \draw[branch_stream] (split2) -- (norm2.west);
        \draw[branch_stream] (norm2.east) -| (mlp.south);
        \draw[branch_stream] (mlp.north) -- node[hook_tri] (mlp_tri) {} (mlp.north |- add2.east) -- (add2.east); 
    
        
        \node[hook, anchor=west] (attn_lbl) at (attn_tri -| label_col) {attn\_out};
        \draw[densely dashed, black!50] (attn_tri.east) -- (attn_lbl.west);
    
        \node[hook, anchor=west] (mlp_lbl) at (mlp_tri -| label_col) {mlp\_out};
        \draw[densely dashed, black!50] (mlp_tri.east) -- (mlp_lbl.west);
    
        \node[hook, anchor=west] (resid_lbl) at (resid_tri -| label_col) {resid\_post};
        \draw[densely dashed, black!50] (resid_tri.east) -- (resid_lbl.west);
    
    \end{tikzpicture}
    }
    \caption{Internal architecture of a transformer block. Each selected hook point acts as a checkpoint on model operations in the model's embedding space. \textit{attn\_out} checks the mixing across the sequence, \textit{mlp\_out} checks the transformation, and \textit{resid\_post} checks the full block operation.}
    \label{fig:transformer_block}
\end{figure}

\subsubsection{Stage 4A: All or Nothing}
We first looked at the performance with none of the internal activations distilled (\textbf{Only Outputs}) and with all of them distilled (\textbf{Balanced}). This allowed us to determine whether the maximum additional information from the internals might have introduced a catastrophic amount of noise. Table \ref{tab:stage4a} shows a fairly distinct advantage for the Balanced objective with an online projector over all other configurations. Since each projector performed identically for the Only Outputs objective, we proceeded with only the randomized projector, as it performs identically to the other projectors without requiring training.

\begin{table}[htbp]
\centering
\begin{tabular}{llc}
\toprule
\textbf{Objective Components} & \textbf{Projection} & \textbf{Accuracy} \\
\midrule
             & Online     & \textbf{0.4160} \\
Balanced     & Offline    & 0.4150 \\
             & Randomized & 0.4150 \\
\midrule
             & Online     & 0.4151 \\
Only Outputs & Offline    & 0.4151 \\
             & Randomized & 0.4151 \\
\bottomrule
\end{tabular}
\caption{Stage 4a results, comparing different projections and objective components.}
\label{tab:stage4a}
\end{table}

\subsubsection{Stage 4B: Marginal Modifications}
In the second part of Stage 4, we made marginal modifications to the previous objective configurations. For the Only Outputs objective with the randomized projector, we explored the isolated effects of including one of the groups of hooks (\textit{embed}, \textit{attn\_out}, \textit{mlp\_out}, \textit{resid\_post}). All four experiments achieved the same performance of 0.4142.

For the Balanced objective with the online projector we took the opposite approach, \textit{removing} one of hook groups from the original objective. The results in Table \ref{tab:stage4b_sub} demonstrate that the Balanced objective is best, but also that the \textit{mlp\_out} activations seem to be the most important of the included components. 

\begin{table}[htbp]
\centering
\begin{tabular}{lc}
\toprule
\textbf{Objective Components} & \textbf{Accuracy} \\
\midrule
No Resid & \textbf{0.4154} \\
No Attn Out & 0.4145 \\
No Embed & 0.4113 \\
No MLP & 0.4040 \\
\bottomrule
\end{tabular}
\caption{Stage 4B subtractive results, removing specific objective components from original Balanced objective.}
\label{tab:stage4b_sub}
\end{table}

\subsection{Stage 5: Final Validation}
At the final stage of our exploration, we revisited the divergence parameter and the projection direction. We did not find the results from the divergence parameter search from Stage 3 exceptionally convincing and thought it was worth exploring further. As mentioned in Section \ref{ss:stage2}, our exploration proceeded with the upward projection direction. Since the online projection was the only variant with any performance deviation when changing direction and ended up being the best choice to this point, we further explore the downward online projection.

With the pretrained student, online projection, and Balanced objective fixed, we once again ran through a sweep of the divergence parameters. Table \ref{tab:stage5_beta} shows how our previous value of 0.25 remains the best option.

\begin{table}[htbp]
\centering
\begin{tabular}{cc}
\toprule
\textbf{$\beta$} & \textbf{Accuracy} \\
\midrule
0.0  & 0.3984 \\
0.25 & \textbf{0.4160} \\
0.5  & 0.4128 \\
0.75 & 0.4136 \\
1.0  & 0.4149 \\
\bottomrule
\end{tabular}
\caption{Stage 5 results revisiting divergence parameter sweep.}
\label{tab:stage5_beta}
\end{table}

Finally, we ran a head-to-head comparison with the upward and downward projection directions. Table \ref{tab:stage5_proj_dir} shows the results indicating that the upward direction performs better. One of the interesting points to note is that the training loss values for upward projections are an order of magnitude higher than those of downward projections. We attribute this to the curse of dimensionality, as the teacher model's embeddings are twice the size of the student's.

\begin{table}[htbp]
\centering
\begin{tabular}{cc}
\toprule
\textbf{Projection Direction} & \textbf{Accuracy} \\
\midrule
Down  & 0.4106 \\
Up & \textbf{0.4160} \\
\bottomrule
\end{tabular}
\caption{Stage 5 results revisiting upward vs. downward projection direction.}
\label{tab:stage5_proj_dir}
\end{table}

\section{Hyperparameters}
As part of our explorations, we used Stage 1 to tune the hyperparameters. With so many interdependencies, we ran 25 trials for each of the training configurations, utilizing early stopping to increase efficiency. Interestingly, there was majority consensus on the hyperparameter values across the different configurations, so the remaining exploration stages and all of the presented results utilized the same final hyperparameter values. Table \ref{tab:tune-hparams} shows the search space for each of the hyperparameters. Table \ref{tab:final-hparams} shows the finalized hyperparameters, which were used to train all students except for the TED models. Table \ref{tab:hparams-off-proj} shows the final hyperparameters used for training the offline projectors. Table \ref{tab:ted_hparams_search} shows the search space and Table \ref{tab:ted_hparams_final} shows the final values for both learning rates in our TED implementation.

\begin{table}[htbp]
\centering
\begin{tabular}{lcc}
\toprule
\textbf{Hyperparameter} & \textbf{Minimum} & \textbf{Maximum} \\
\midrule
Learning Rate           & $1 \times 10^{-6}$ & $1 \times 10^{-3}$ \\
Weight Decay            & $1 \times 10^{-4}$ & 0.1                \\
Temperature             & 0.1                & 2.0                \\
Max Grad Norm           & 0.1                & 2.0                \\
\bottomrule
\end{tabular}
\caption{Hyperparameter Search Space}
\label{tab:tune-hparams}
\end{table}

\begin{table}[htbp]
\centering
\small
\begin{tabular}{lcccc}
\toprule
\textbf{Model (Training)} & \textbf{Learning Rate} & \textbf{Weight Decay} & \textbf{Max Grad Norm} & \textbf{Temperature} \\
\midrule
Teacher (FT)               & $2 \times 10^{-5}$    & $2 \times 10^{-4}$    & 0.5                   & --                  \\
\midrule
Pretrained (FT)        & $3.4 \times 10^{-5}$  & $1 \times 10^{-2}$    & 2.0                   & 1.0                 \\
Pretrained (KD)        & $2.5 \times 10^{-5}$  & $4.9 \times 10^{-4}$  & 0.4                   & 0.8                 \\
\midrule
Shallow (FT)           & $4.9 \times 10^{-5}$  & $3 \times 10^{-4}$    & 0.4                   & 0.3                 \\
Shallow (KD)           & $4.3 \times 10^{-4}$  & $4.9 \times 10^{-2}$  & 0.3                   & 0.5                 \\
\bottomrule
\end{tabular}
\caption{Final Hyperparameter Settings for Model Training. \textit{FT} indicates the parameters used for finetuning, and \textit{KD} indicates the parameters used for all distillation methods.}
\label{tab:final-hparams}
\end{table}

\begin{table}[h]
\centering
\begin{tabular}{lc}
\toprule
\textbf{Hyperparameter} & \textbf{Value} \\
\midrule
Learning Rate           & $4.5 \times 10^{-4}$ \\
Weight Decay            & $1.3 \times 10^{-2}$ \\
Tied Weights            & True                 \\
\bottomrule
\end{tabular}
\caption{Final Hyperparameters for Offline Projectors}
\label{tab:hparams-off-proj}
\end{table}

\begin{table}[htbp]
    \centering
    \begin{tabular}{lcc}
        \toprule
         \textbf{Hyperparameter} & \textbf{Min} & \textbf{Max} \\
         \midrule
         LR (filters) & 1e-4 & 1e-2 \\
         LR (distillation) & 1e-6 & 1e-4 \\ 
         \bottomrule
    \end{tabular}
    \caption{The search space for the hyperparameters in TED. Just as in the rest of our tuning, we searched this space over 25 trials.}
    \label{tab:ted_hparams_search}
\end{table}

\begin{table}[htbp]
    \centering
    \begin{tabular}{lc}
        \toprule
         \textbf{Hyperparameter} & \textbf{Value} \\
         \midrule
         LR (filters) & 3.9e-4 \\
         LR (distillation) & 3.66e-5 \\ 
         \bottomrule
    \end{tabular}
    \caption{The final hyperparameter settings used for TED training.}
    \label{tab:ted_hparams_final}
\end{table}

\end{document}